\documentclass[lettersize,journal]{IEEEtran}
\usepackage{amsmath,amsfonts,amssymb,bbm}
\usepackage{algorithmic}
\usepackage{algorithm}
\usepackage{array}
\usepackage[caption=false,font=footnotesize,labelfont=rm,textfont=rm]{subfig}
\usepackage{textcomp}
\usepackage{stfloats}
\usepackage{url}
\usepackage{verbatim}
\usepackage{graphicx}
\usepackage{cite}
\usepackage{booktabs}      
\usepackage{nicefrac}      
\usepackage{microtype}      
\usepackage[dvipsnames, svgnames, x11names]{xcolor}        
\usepackage{bm}             
\usepackage{enumitem}       
\usepackage{multirow}
\usepackage{diagbox}
\usepackage{colortbl}       
\definecolor{mygray}{gray}{.9}
\usepackage{bbding}
\usepackage{hyperref}    

\definecolor{lightgray}{gray}{0.5}
\begin{document}

\title{Disjoint Masking with Joint Distillation for \\ Efficient Masked Image Modeling}

\author{Xin Ma, 
        Chang Liu, 
        Chunyu Xie, 
        Long Ye, 
        Yafeng Deng, 
        and Xiangyang Ji
\thanks{X. Ma and L. Ye are with the School of Information and Communication Engineering, Communication University of China, Beijing 100024, China, Email: \{mx\_mark, yelong\}@cuc.edu.cn.}
\thanks{C. Liu and X. Ji are with the Department of Automation, Tsinghua University, Beijing 100084, China, Email: \{liuchang2022, xyji\}@tsinghua.edu.cn.} 
\thanks{C. Xie and Y. Deng are with 360 AI Research, Beijing, China, Email: yuxie@buaa.edu.cn, dengyafeng@gmail.com.}
\thanks{X. Ji is the corresponding author.}
}

\maketitle
 
\begin{abstract}
Masked image modeling (MIM) has shown great promise for self-supervised learning (SSL) yet been criticized for learning inefficiency.
We believe the insufficient utilization of training signals should be responsible.
To alleviate this issue, we introduce a conceptually simple yet learning-efficient MIM training scheme, termed \textit{D}isjoint \textit{M}asking with \textit{J}oint \textit{D}istillation (DMJD).
For disjoint masking (DM), we sequentially sample multiple masked views per image in a mini-batch with the disjoint regulation to raise the usage of tokens for reconstruction in each image while keeping the masking rate of each view.
For joint distillation (JD), we adopt a dual branch architecture to respectively predict invisible (masked) and visible (unmasked) tokens with superior learning targets.
Rooting in orthogonal perspectives for training efficiency improvement, DM and JD cooperatively accelerate the training convergence yet not sacrificing the model generalization ability.
Concretely, DM can train ViT with half of the effective training epochs ($3.7\times$ less time-consuming) to report competitive performance. With JD, our DMJD clearly improves the linear probing classification accuracy over ConvMAE by 5.8$\%$. 
On fine-grained downstream tasks like semantic segmentation, object detection, \textit{etc.}, our DMJD also presents superior generalization compared with state-of-the-art SSL methods.
The code and model will be made public at \href{https://github.com/mx-mark/DMJD}{https://github.com/mx-mark/DMJD}.
\end{abstract}


\section{Introduction}
\label{sec:introduction}
\IEEEPARstart{W}{ith} the surge of vision transformers~\cite{2017arXiv170603762V, 2021arXiv210313915S, liu2021swin, Touvron2021TrainingDI, Touvron2021GoingDW, yuan2021tokens}, masked image modeling (MIM) has recently prevailed on the self-supervised representation learning (SSL) leaderboard~\cite{caron2021emerging,chen2021empirical,2021arXiv210609785L}. These approaches assimilate context semantics by predicting a portion of masked tokens. It has been justified that discrete visual tokens~\cite{2021arXiv210608254B, 2021arXiv211112710D, 2021arXiv211210740E, 2021arXiv211107832Z}, raw pixels~\cite{2021arXiv211106377H, 2021arXiv211109886X}, and hand-crafted features~\cite{2021arXiv211209133W} are suitable targets to learn versatile models for a broad spectrum of downstream tasks.

However, MIM methods typically require tremendous training costs, \textit{eg.}, thousands of pre-training epochs, which overburdens academic and industrial communities. In general, an input image is masked out with a high masking rate, also named corruption rate ($m_{corr}$). While the rest unmasked patches are ignored for self-supervision. We argue that a large proportion of input signals is not fully exploited for model learning at each loop mainly accounts for the training inefficiency. Actually, the training efficiency of MIM is largely determined by the prediction rate ($m_{pred}$), the overall portion of input images used to provide supervision for invisible region reconstruction. 
\begin{figure}[!t]
\centering
\subfloat[Varying both ($m_{corr}$ = $m_{pred}$).]{
    \begin{minipage}{0.23\textwidth}
      \includegraphics[width=\linewidth]{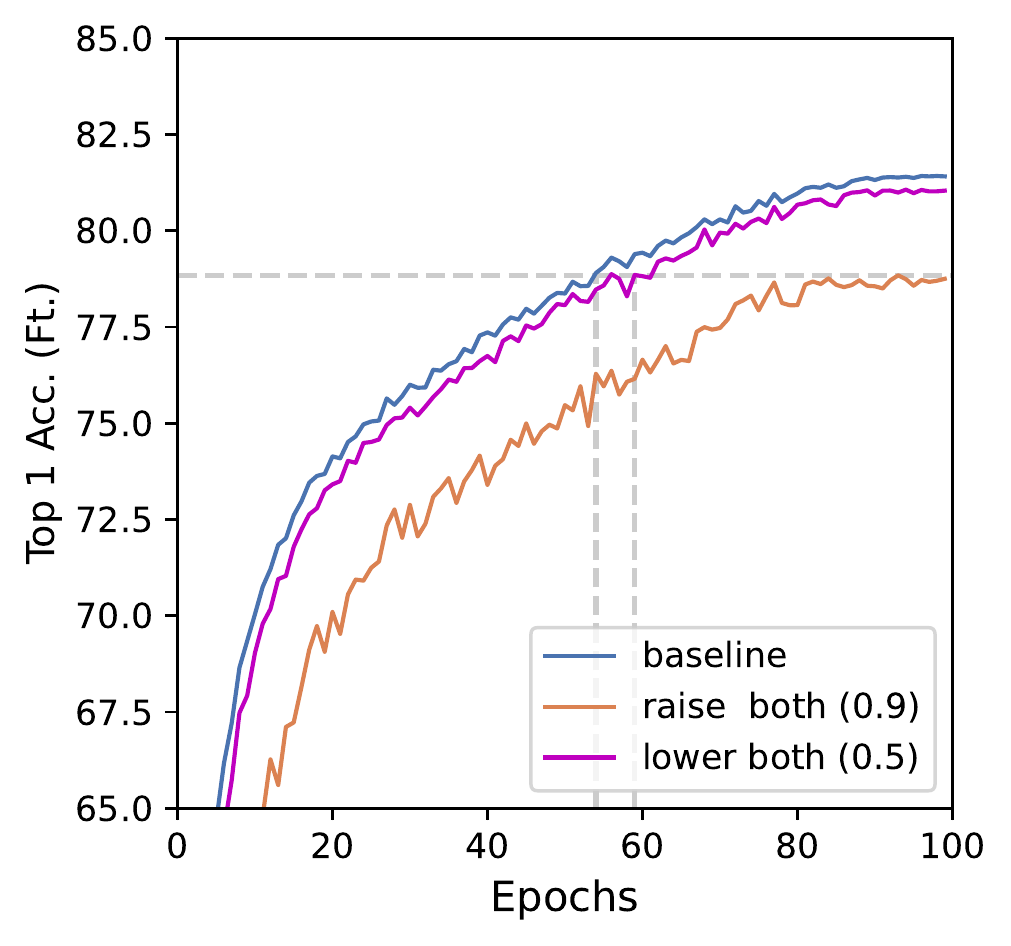}
        \label{fig1:a}
        \vspace{-4mm}
    \end{minipage}
}
\subfloat[Varying $m_{pred}$ ($m_{corr}=0.75$).]{
    \begin{minipage}{0.23\textwidth}
      \includegraphics[width=\linewidth]{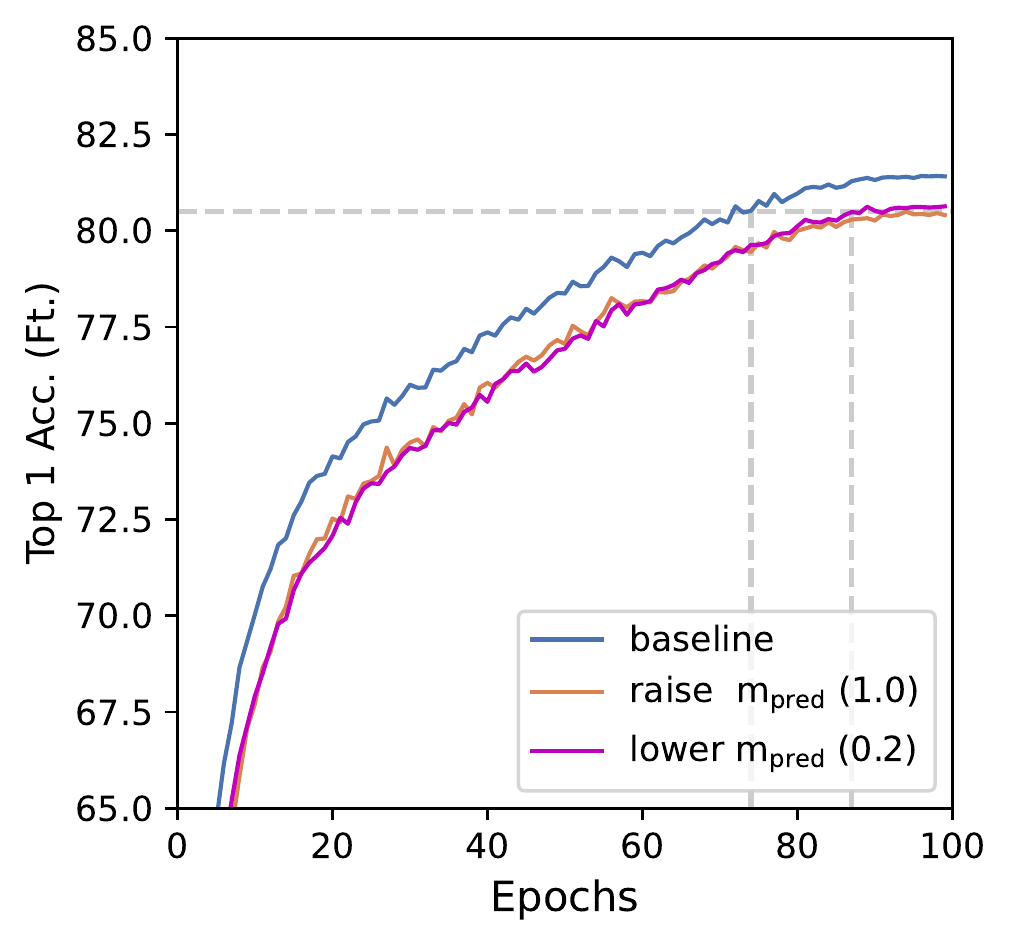}
        \label{fig1:b}
        \vspace{-4mm}
    \end{minipage}
}
\quad
\subfloat[Varying $m_{corr}$ ($m_{pred}=0.75$).]{
    \begin{minipage}{0.23\textwidth}
      \includegraphics[width=\linewidth]{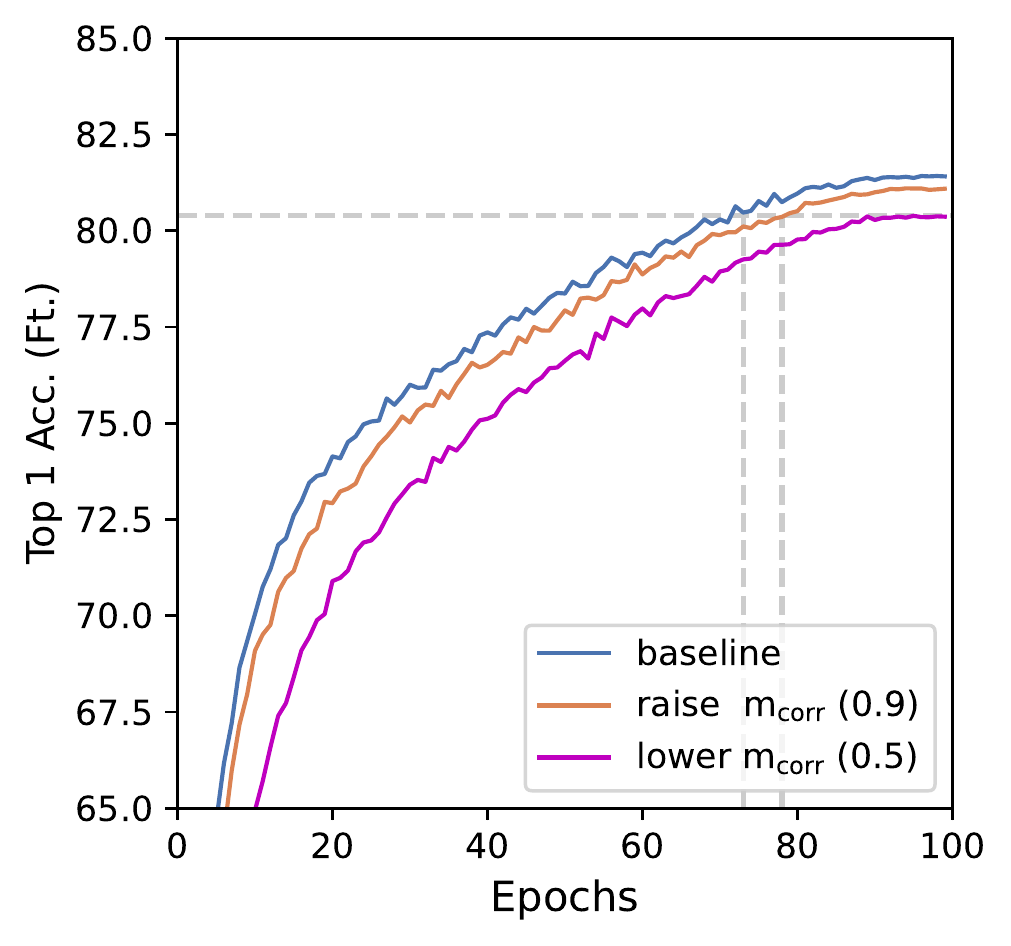}
        \label{fig1:c}
        \vspace{-4mm}
    \end{minipage}
}
\subfloat[DMJD (Ours).]{
    \begin{minipage}{0.23\textwidth}
      \includegraphics[width=\linewidth]{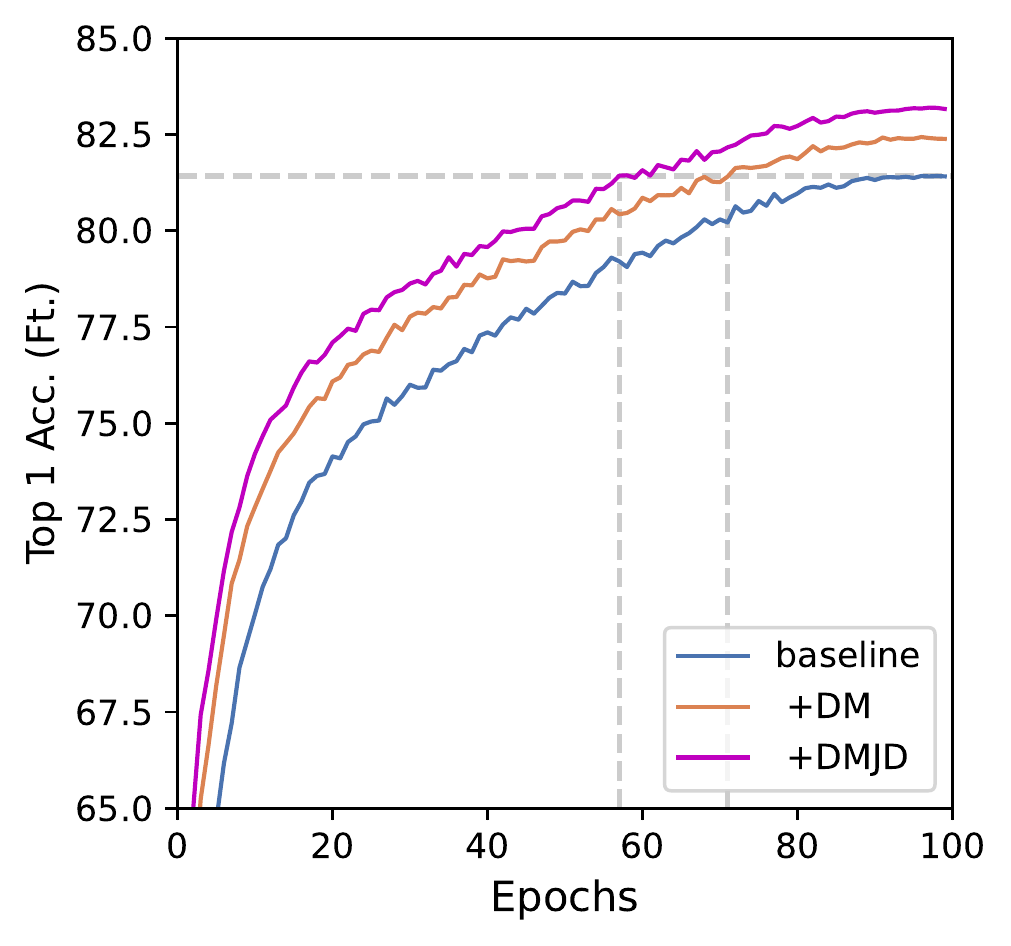}
        \label{fig1:d}
        \vspace{-4mm}
    \end{minipage}
}
\vspace{4mm}
\caption{MIM Training efficiency analysis. 
Specifically, we pre-train models based on MAE~\cite{2021arXiv211106377H} for 100 epochs on ImageNet-1k~\cite{Deng2009ImageNetAL} with uniform masking and plot the corresponding convergence curve on the validation set during fine-tuning. The blue curve in each sub-figure denoted as the ``baseline" is reported under the optimal setting of $m_{pred}=m_{corr}=75\%$.
In sub-figures (a), (b), and (c), no matter how $m_{corr}$ and $m_{pred}$ change in each view of an image, the optimal choice for fast convergence is the baseline setting. It inspires us to develop a multi-view masking strategy, \textit{i.e.}, DM, to raise the overall $m_{pred}$ in an image while keeping $m_{pred}$ equal to $m_{corr}$ in each view. In the spirit of increasing the utilization of input signals, we also introduce JD with an additional visible distillation branch, which works cooperatively with DM to expedite the training procedure, sub-figure (d).}
\label{fig:motivation}
\end{figure}

\begin{figure}[!t]
\centering
    \subfloat[Vanilla MIM.]{
        \begin{minipage}{0.48\textwidth}
          \includegraphics[width=\linewidth]{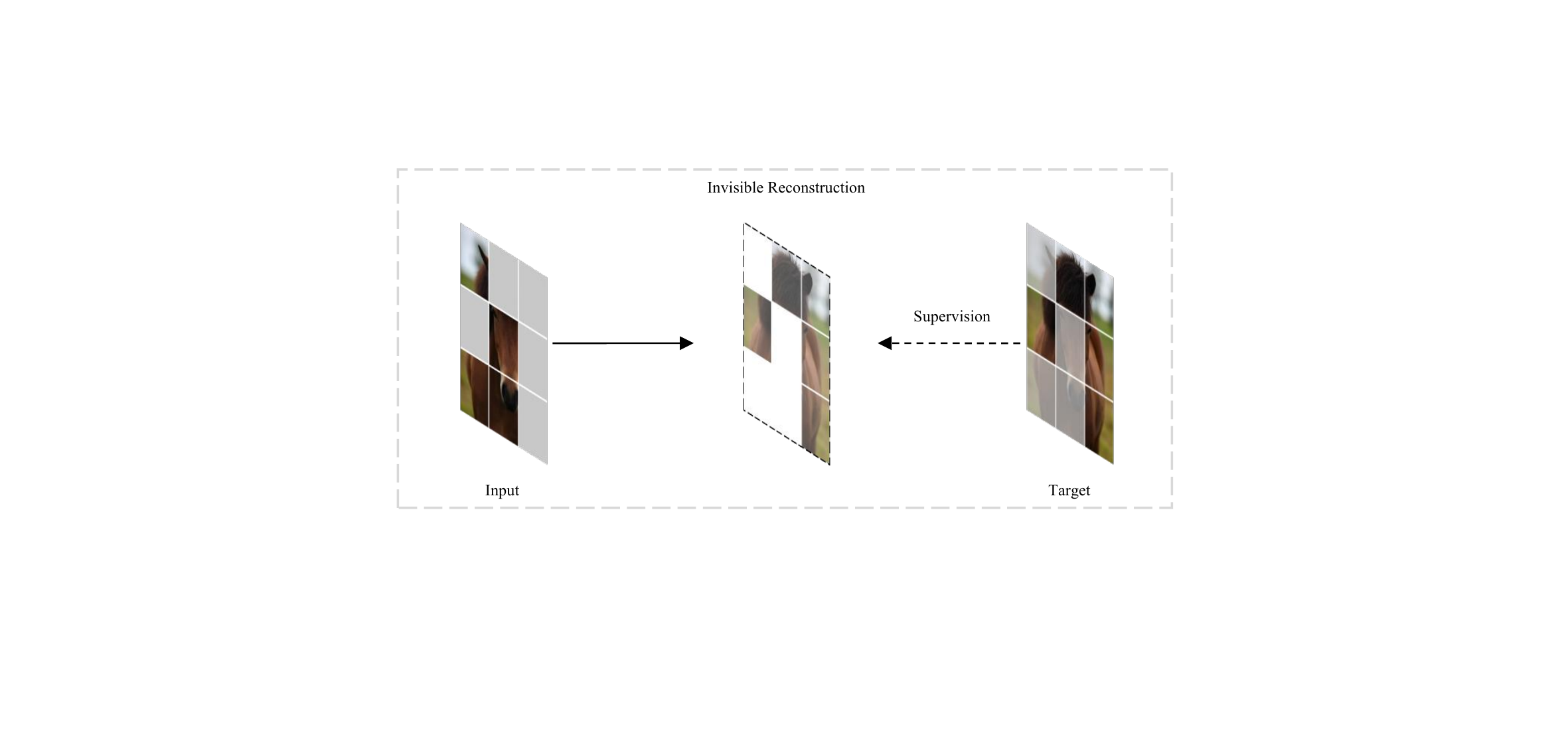}
        \end{minipage}
    }
    \quad
    \subfloat[DMJD (Ours).]{
        \begin{minipage}{0.48\textwidth}
          \includegraphics[width=\linewidth]{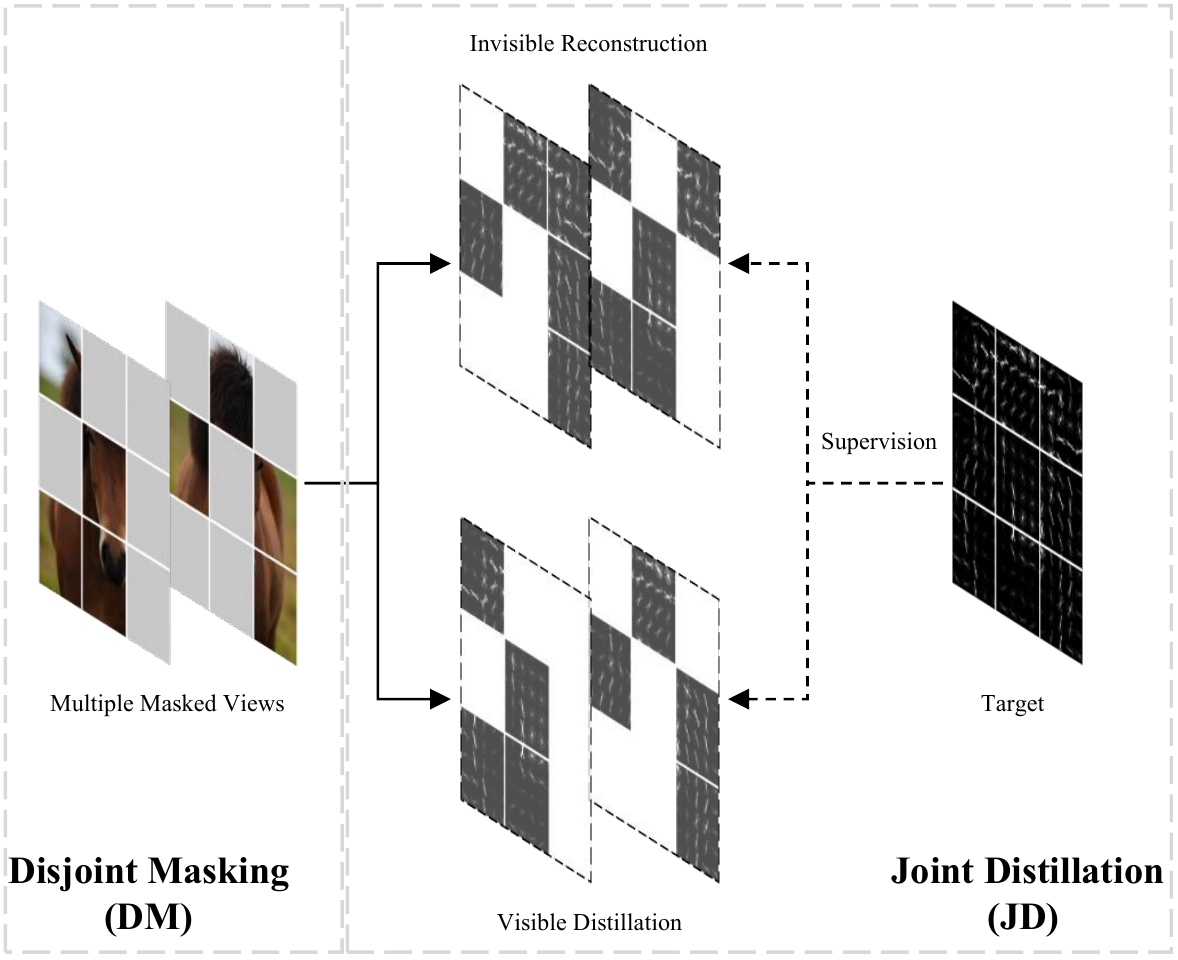}
        \end{minipage}
    }
\vspace{4mm}
\caption{Pipeline illustration of vanilla MIM and our DMJD.
Vanilla MIM methods typically adopt a single-view masking strategy to perform invisible reconstruction with $m_{corr} = m_{pred}$. In contrast, our DMJD introduces a multi-view masking strategy, \textit{i.e.}, DM, to increase the overall prediction rate of each image while keeping $m_{corr} = m_{pred}$ in each view and a dual-branch joint distillation architecture with an additional visible distillation branch to take full use of the input signals with superior targets, \textit{eg.} HOG. 
}
\label{fig:viz}
\end{figure}
To go deeper into how $m_{corr}$ and $m_{pred}$ affect MIM training efficiency, we quantitatively visualize the model convergence curve with various settings in Fig.~\ref{fig:motivation}. 
Concretely, in Fig.~\ref{fig1:a} and \ref{fig1:c}, one can see that only with a proper corruption rate, \textit{i.e.}, 0.75 with uniform masking, MIM exhibits superior effectiveness and efficiency.
This is because since $m_{corr}$ defines the difficulty of the reconstruction task, the training efficiency intrinsically depends on $m_{corr}$. With a too-high corruption rate, there is not enough context to correctly recover the masked patches~\cite{2021arXiv211106377H}, making the model more intricate and unpredictable. And a low corruption rate will degrade the model performance because the model can trivially recover a missing patch from neighboring patches without needing a high-level visual understanding. 
What is worse, with the optimal setting of $m_{corr}= 0.75$, naively varying $m_{pred}$ also hampers the model training, Fig.~\ref{fig1:b}.
These facts reveal the challenges of improving MIM training efficiency and inspire us to think differently to increase the utilization of input signals.

In this paper, we develop a conceptually simple yet learning-efficient MIM training scheme, termed Disjoint Masking with
Joint Distillation (DMJD), Fig.~\ref{fig:viz}. 
Disjoint masking (DM) is a multiple-view sampling strategy targeting flexibly raising the prediction rate of each input while keeping the corruption rate of each view. 
Concretely, we sequentially sample a series of masked views of an image from its disjoint portions, \textit{i.e.}, overall have been previously unmasked/masked patches, for wider coverage of the invisible regions in a training loop for reconstruction.
Since the increasing utilization of training signals in a batch may reduce the gradient variance and hamper the generalization of learned models~\cite{2016arXiv160904836S, Hoffer2020AugmentYB, 2021arXiv210513343F}, we introduce an adaptive learning rate scale rule taking the relative prediction rate increment as a scale factor to enhance the model training. 

Joint distillation (JD) performs distillation on visible and invisible regions with a dual branch architecture to further increase training efficiency.
In addition to the original masked prediction branch (MPB), we add a visible distillation branch (VDB) to provide semantic guidance through parametric or non-parametric tokenizers for visible token learning.
With increased prediction rates, visible distillation, and superior targets, the proposed DMJD ensures higher learning efficiency from orthogonal perspectives yet works cooperatively.

It is worth noting that our DMJD accelerates training convergence but not sacrificing model generalizability. 
Specifically, ViT/ConViT~\cite{2021arXiv210313915S, 2022arXiv220503892G} equipped with DMJD typically improves performances with improved training efficiency not only on ImageNet-1K classification but also on fine-grained downstream tasks, \textit{eg.} semantic/instance segmentation, object detection, \textit{etc.}
Take an example, for linear probing classification on MaskFeat~\cite{2021arXiv211209133W} and ConvMAE~\cite{2022arXiv220503892G} baselines, DMJD achieves performance gains of 3.4$\%$ and 5.8$\%$ with $1.8 \times$ and $3 \times$ acceleration.
We hope these observations shed new light on MIM training efficiency research in the community.

Our contributions can be summarized as follows: 
\begin{itemize}[itemsep=2pt,topsep=0pt,parsep=0pt]
\item  We propose a conceptually simple yet learning-efficient MIM training scheme, termed disjoint masking with joint distillation (DMJD), which targets increasing the utilization of per image at each training loop.

\item We devise a multi-view generation strategy, \textit{i.e.}, disjoint masking (DM), to increase the prediction rate while keeping the corruption rate for efficient MIM and introduce the adaptive learning rate scale rule for better model generalization with augmented training batches.

\item We develop a dual-branch architecture for joint distillation (JD), effectively pursuing representation learning on both visible and invisible regions with superior targets.

\item We conduct sufficient evaluations justifying our DMJD can significantly accelerate model convergence and achieve outstanding performances on standard benchmarks.
\end{itemize}

\section{Related Work}
\subsection{Masked Image Modeling}
Since ViT~\cite{2021arXiv210313915S} overcomes the architectural obstacle of applying masked signal modeling for visual pre-training, MIM has achieved great success recently. 
The increasing attention can be roughly categorized into three aspects, including learning targets, backbone extensions, and masking strategies.  

Learning targets are critical since they provide explicit guidance for visual learning through reconstruction 
and their characteristics can be injected into the learned model. 
MAE~\cite{2021arXiv211106377H} and SimMIM~\cite{2021arXiv211109886X} reveal that raw pixels as reconstruction targets are adequate for effective visual learning. MaskFeat~\cite{2021arXiv211209133W} introduces local invariant features, such as HOG~\cite{Dalal2005HistogramsOO}, to avoid modeling high-frequency low-level details in the pixel space and concentrate meaningful semantic abstraction.
BEiT~\cite{2021arXiv210608254B} and PeCo~\cite{2021arXiv211112710D} further apply a discrete visual codebook produced by dVAE~\cite{2021arXiv210212092R} with an additional pre-training stage for high-level semantic abstraction. 
To get rid of extra pre-training, iBOT~\cite{2021arXiv211107832Z} and data2vec~\cite{Baevski2022data2vecAG} jointly optimize the model and the target tokenizer.
With the fast development of multi-modal foundation models, CLIP~\cite{Radford2021LearningTV} is actively exploited yet not limited as an effective target tokenizer for MIM~\cite{2022arXiv220305175W, 2022arXiv220806366P, 2022arXiv220810442W, 2022arXiv220806049H, 2022arXiv221109799Z}.

As advanced backbones take advantage of more discriminative representations, researchers are inspired to extend MIM to multi-scale hybrid convolution-transformer architectures~\cite{2021arXiv211109886X, 2022arXiv220503892G}.
Also, to explore what kind of information in images to be predicted benefits the model learning, several masking strategies~\cite{Li2021MSTMS, 2022arXiv220113100S, 2022arXiv220312719K, 2022arXiv220610207L} have been proposed to improve the vanilla random masking.
However, tremendous training costs for convergence limit MIM methods for application and research. The training efficiency issue becomes urgent to be attended.

\begin{figure*}[t]
  \centering
  \includegraphics[width=1\textwidth]{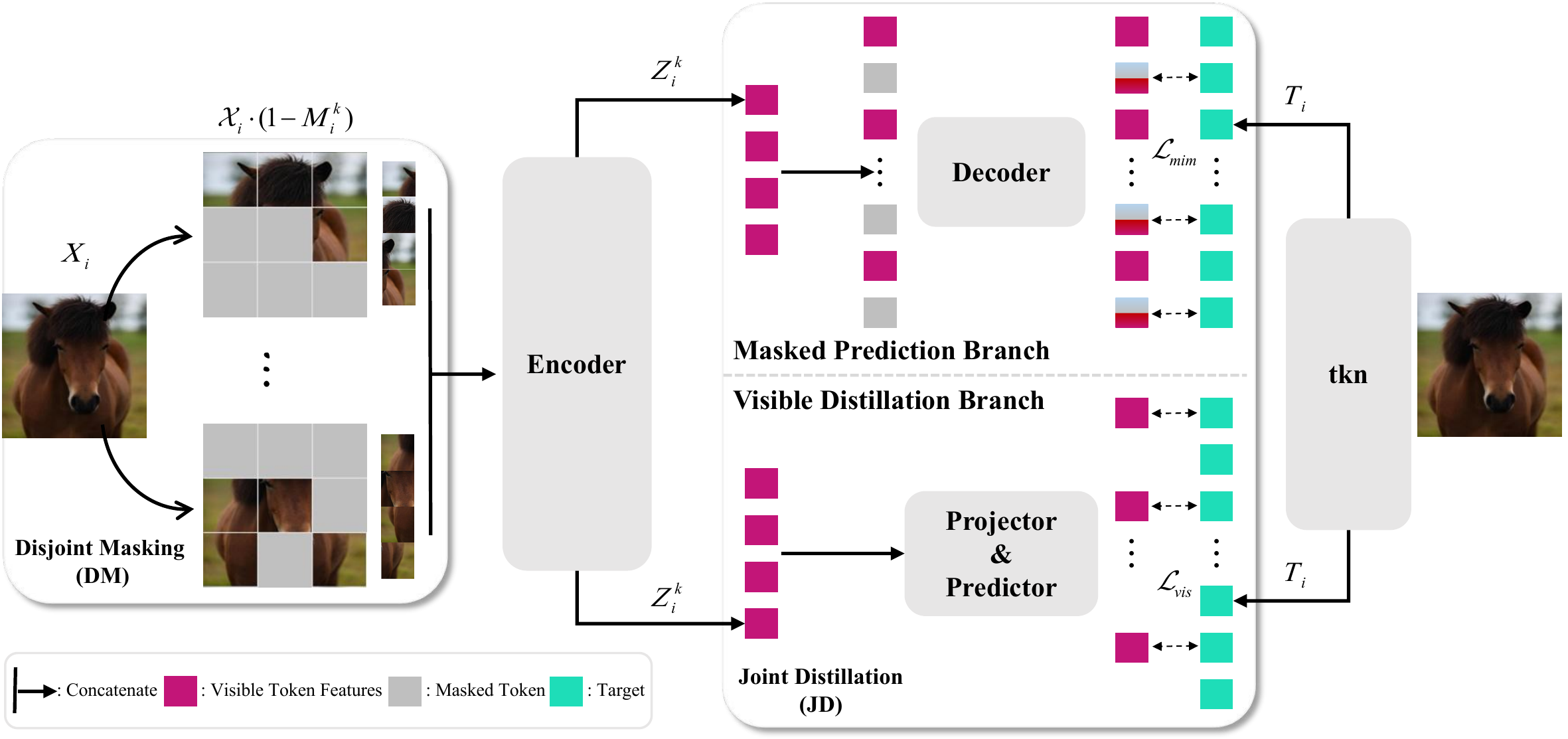}
  \caption{The pipeline of the proposed DMJD. During pre-training, $K$ masked views of each image are randomly sampled in a mini-batch with DM. Then, they will be fed to the encoder and the dual branch decoder for invisible reconstruction and visible distillation with targets extracted by the tokenizer $tkn(\cdot)$.}
  \label{fig:arch}
\end{figure*}

\subsection{MIM Training Efficiency}
Powerful computation resources have grown rapidly to facilitate fast training with lower numerical precision and larger batch sizes for large-scale models~\cite{2017arXiv170602677G, 2017arXiv170508741H, 2018arXiv180711205J, 2018arXiv181106992Y, 2017arXiv170803888Y, Hoffer2020AugmentYB, 2019arXiv190205509B, 2021arXiv210513343F, 2019arXiv190705550C}. 
However, the high computation and time costs are still an obstacle to the practical applications of MIM.   
A feasible attempt is to apply efficient hierarchical transformers. Concretely, UM-MAE~\cite{2022arXiv220510063L} designs a masking strategy to preserve the relative position relationship between visible patches. 
GreenMIM~\cite{2022arXiv220513515H} proposes a uniform partition to ensure that visible patches within each local window can be grouped with equal size. 
MixMIM~\cite{2022arXiv220513137L} creates a mixed view that the masked tokens from one image are replaced with visible tokens from another image.
An alternate way for MIM training efficiency is reducing the computational complexity of invisible reconstruction. 
Specifically, LoMaR~\cite{2022arXiv220600790C} performs masked reconstruction within small 7×7 windows. 
FastMIM~\cite{2022arXiv221206593G} directly pre-trains model with low-resolution inputs.

Differently, inspired by MLM works~\cite{2020arXiv200310555C,2022arXiv220208005W} which attempt to supervise a larger portion of input signals for learning efficiency, our DMJD is devised to increase the utilization of each image with a multi-view masking strategy~\cite{8695120,9548841,8059841,9127152} and dual-branch distillation using high-level learning targets.

\section{Method}
\subsection{Overview}
MIM methods are typically built on ViTs~\cite{2021arXiv210313915S}, where an input image $\bm{X}_{i}$ will be split into a set of non-overlapping patches and linearly projected to a sequence of tokens. This process can be formulated as $\mathcal{X}_{i} = \{(\bm{x}_{ij}, \bm{p}_{ij})_{j=1}^N\}$, where $\bm{p}_{ij}$ denotes the positional embedding of the $j$-th token $\bm{x}_{ij}$. To yield masked views for reconstruction, specific masking strategies can be defined as
\begin{equation}
    \mathbb{M}(\mathcal{X}_{i}, m_{patt}, m_{corr}) = \bm{M}_{i},
    \label{eq:masking function}
\end{equation}
where $\bm{M}_{i} = (m_{ij})_{j=1}^{N}$, $m_{ij}$ equals 1 if the $j$-th token $\bm{x}_{ij}$ needs to be masked, otherwise 0. $m_{corr}$ and $m_{patt}$ are the corruption rate and the masking pattern, \textit{eg.} random, block-wise, \textit{etc.}
The reconstruction loss function is defined as
\begin{equation}
    \mathcal{L}_{mim} = -\sum_{i = 1}^{B} \sum_{j = 1}^{N} m_{ij} \cdot \ell (\bm{x}'_{ij}, \bm{t}_{ij}),
    \label{eq:mim_loss}
\end{equation}
where $\bm{x}'_{ij}$ and $\bm{t}_{ij}$ are the network prediction and the corresponding reconstruction target of $\bm{x}_{ij}$. $B$ is the batch size. $\ell (\cdot)$ is typically the mean squared error (MSE).

Since gradients are only back-propagated on masked tokens, a small subset of an image, it may cause the training inefficiency problem~\cite{2020arXiv200310555C,2022arXiv220208005W}.
And it is not wise to simply increase the masking rate or recover unmasked patches, which even slows down the convergence, Fig.~\ref{fig:motivation}. Thus, we propose disjoint masking with joint distillation (DMJD), a dual-branch distillation framework with a multiple-masked view sampling approach to increase the prediction rate with a fixed corruption rate for efficient MIM, Fig.~\ref{fig:arch}. 

Concretely, we impose a masking regulation to generate multiple complementary views facilitating more invisible tokens of each image to be reconstructed in the masked prediction branch (MPB). 
In the visible distillation branch (VDB), we bridge the semantic gap between the visible token features and their corresponding learning targets through distillation, further enhancing the training efficiency~\cite{2021arXiv211209133W, Baevski2022data2vecAG, 2022arXiv220514141W}. 

\subsection{Disjoint Masking}

\label{sec:dm}
\begin{figure*}[!t]
  \centering
  \includegraphics[width=\textwidth]{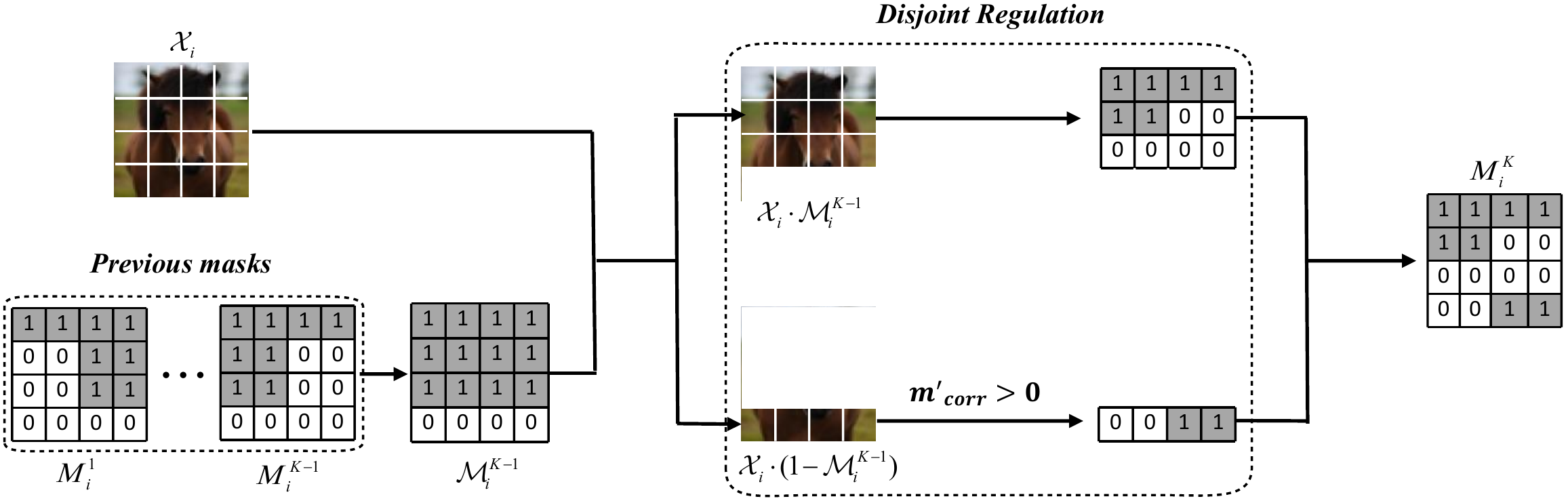}
  \caption{Illustration of disjoint masking. With DM, we sequentially sample the $K$-th view from disjoint regions of the currently have not been masked token set $\mathcal{X}_{i} \cdot \mathcal{M}_{i}^{K-1}$ and the rest part $\mathcal{X}_{i} \cdot (\bm{1}-\mathcal{M}_{i}^{K-1})$ with the disjoint regulation, \textit{i.e.}, keeping the corruption rate $m^{'}_{corr}$ in $\mathcal{X}_{i} \cdot (\bm{1}-\mathcal{M}_{i}^{K-1})$ positive. The regulation ensures the overall prediction rate of each image is larger than the pre-defined corruption rate.}
  \label{fig:dm}
\end{figure*}
DM targets sequentially sampling multiple masked views under a regulation to increase the overall portions of tokens used for reconstruction. Since augmented views reduce the samples with a given batch size, We modify the learning rate scale rule to optimize the model training for generalization.

\subsubsection{Disjoint Regulation}
Suppose $\mathcal{M}_{i}^{1}$ = $\mathbb{M}$($\mathcal{X}_{i}$, $m_{patt}$, $m_{corr}$). The set of tokens that have been masked in the previous $K$-1 masks is defined as $\mathcal{M}_{i}^{K-1} = \bm{1}- \prod \limits_{k=1}^{K-1} (\bm{1}-\bm{M}_{i}^{k})$, where $K \geq 2$. Thereafter, DM ($\mathbb{M}_{dm}$) with the disjoint regulation can be formulated based on Eq.~\ref{eq:masking function} as,
\begin{equation}
    \begin{aligned}
        & \mathbb{M}_{dm}(\mathcal{X}_{i}, m_{patt}, m_{corr}, \mathcal{M}_{i}^{K-1})&=& \ \bm{M}_{i}^K, \\
     \text{s.t.}\ &\ \ \ \ \ \ \ \ \ \ \ \ \ \bm{M}_{i}^{K} \cdot (\bm{1} - \mathcal{M}_{i}^{K-1}) &\ne& \ \bm{0}.
    \end{aligned}
    \label{eq:dm}
\end{equation}

It is not hard to find that only if $\bm{M}_{i}^K$ can mask tokens that have never been masked once, Eq.~\ref{eq:dm} can hold. 
Specifically, to sample the $K$-th mask under the regulation, we first split the image tokens into two disjoint subsets: $\mathcal{X}_{i} \cdot \mathcal{M}_{i}^{K-1}$ (previously masked tokens) and $\mathcal{X}_{i} \cdot (\bm{1}-\mathcal{M}_{i}^{K-1})$ (currently unmasked tokens), Fig.~\ref{fig:dm}.
Then, we partially sample tokens to be masked from the unmasked token set with a positive corruption rate $m^{'}_{corr}>0$ and the rest from the masked set.
In this way, DM guarantees that each masked view will include new tokens beyond previous views with a fixed corruption rate while increasing the overall prediction rate of each image. 

\subsubsection{Adaptive Learning Rate Scale}
Since gradients are averaged in a mini-batch, the learning rate scale rule is proposed to alleviate the generalization degradation caused by the over-smoothed gradient variance with large batch sizes~\cite{2017arXiv170602677G}. It normally scales up the base learning rate $\eta_{base}$ with the batch size $b$ as $\eta = \eta_{base} \cdot b/256$, where $256$ is a scale factor. 

With DM, a mini-batch only contains $b_u$ unique samples, where $b_u = b/K$. Since repetitively utilizing each image has a similar effect of enlarging the batch size on the gradient variance, we intend to scale up the learning rate by the relative utilization improvement of tokens used for reconstruction, \textit{i.e.}, $\frac{m_{pred}}{m_{corr}}$.
 Thus, the modified adaptive learning rate scale can be defined as
\begin{equation}
  \eta = \eta_{base} \cdot \left(b_u \cdot \frac{m_{pred}}{m_{corr}}\right) /\ 256.
\end{equation}

\subsection{Joint Distillation}
As shown in Fig.~\ref{fig:arch}, we add an additional visible distillation branch (VDB) to bridge the visible encoding features and its corresponding targets~\cite{2021arXiv211209133W, Baevski2022data2vecAG, 2022arXiv220514141W}.
\subsubsection{Model Architecture}
\paragraph{Projector ($\bm{g}$) \& Predictor ($\bm{q}$)}
We introduce a nonlinear projector to reduce the information loss~\cite{2020arXiv200205709C}.
Specifically, it has a 3-layer MLP-like structure with a LayerNorm (LN)~\cite{2016arXiv160706450L} applied to each fully-connected (FC) layer following SimCLR~\cite{2020arXiv200205709C}. The output dimension of all MLP hidden layers is 512. Only the visible token features are fed to the projector.
To align the output dimensions of the projector and the target tokenizer, we insert a single FC layer as the predictor.
\paragraph{Target Encoding ($tkn(\cdot)$)}
We evaluate two types of targets to introduce direct guidance for visible token learning: i) representative local invariant features, \textit{eg.} HOG~\cite{Dalal2005HistogramsOO}; ii) features extracted by deep models which extract context information well, \textit{eg.} CLIP~\cite{Radford2021LearningTV}. The target encoding process can be expressed as $\bm{T}_i=tkn(\mathcal{X}_i)$.

\subsubsection{Learning Objectives}
We distill the target through Smooth L1 loss in VDB, as
\begin{equation}
\mathcal{L}_{vis}=\left\{
    \begin{array}{ll}
        \frac{1}{2} D^2/\beta ,&  if \left\lvert D\right\rvert \leq  \beta ,\\
        (\left\lvert D\right\rvert -  \frac{1}{2}\beta ) ,& otherwise,
    \end{array}
    \right.
\end{equation}
where $D=\bm{q}(\bm{g}(\bm{Z}_{i}^{k}))-norm(\bm{T}_i \cdot (\bm{1}- \bm{M}_{i}^k))$; $norm(\cdot)$ is a normalization function, \textit{eg.} LN, which enhances the patch-level local contrast for better performance~\cite{2021arXiv211106377H}; $\bm{Z}_{i}^{k}$ and $\bm{T}_i \cdot (\bm{1}- \bm{M}_{i}^k)$ are the encoder outputs of visible tokens and the corresponding targets, respectively;  $\beta$ is experimentally set to 2.0.

Combined with the reconstruction loss $\mathcal{L}_{mim}$ in MPB, Eq.~\ref{eq:mim_loss}, the overall learning objective is updated as,
\begin{equation}
\mathcal{L} = \mathcal{L}_{vis} + \lambda \cdot \mathcal{L}_{mim},
\end{equation}
where $\lambda$ is set to 1 by default. 

\section{Experiment}

\begin{figure}[!t]
\centering
    \includegraphics[width = 0.48\textwidth]{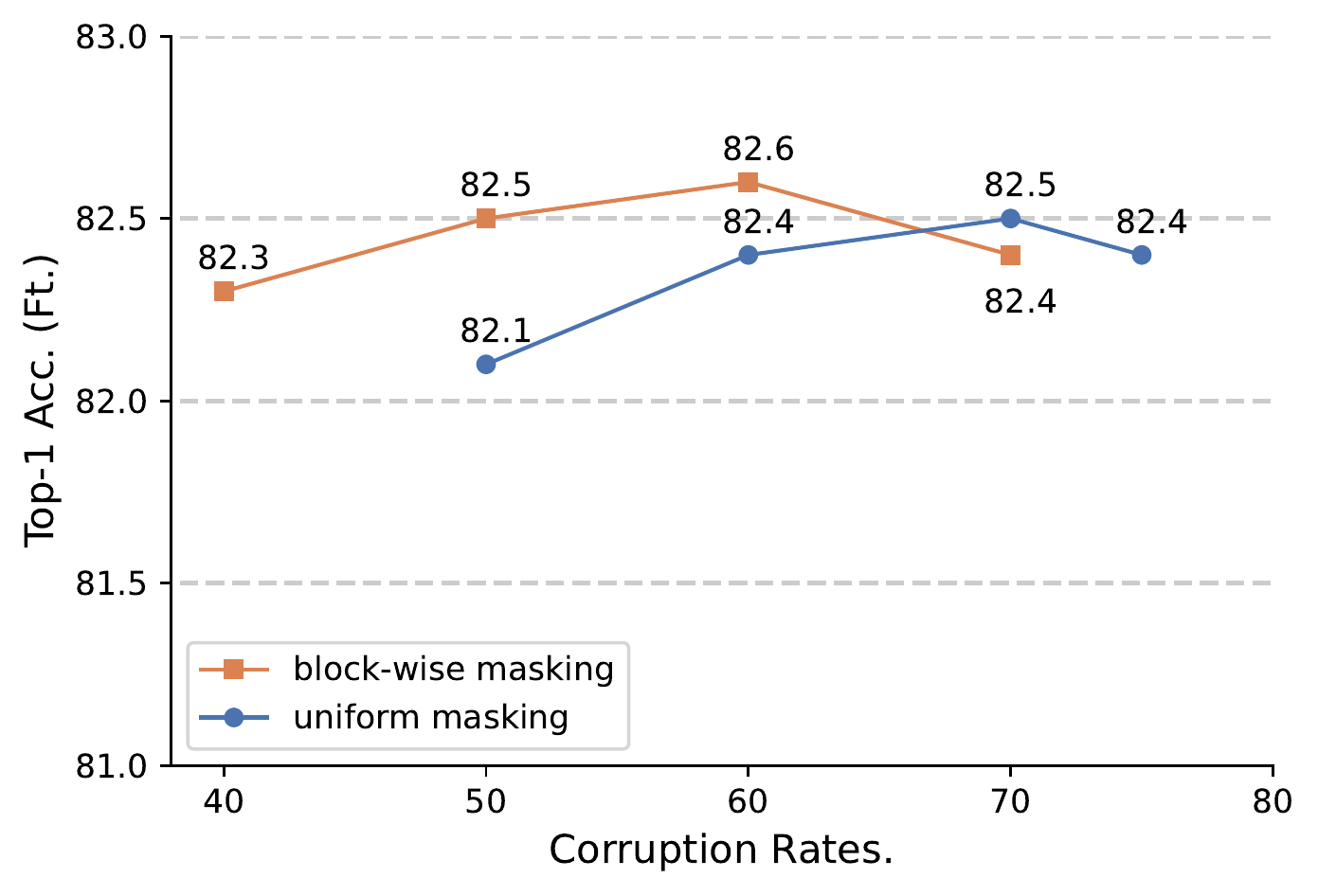}
    \caption{Ablation study about corruption rates under different masking patterns. Fixing the prediction rate at 100$\%$, different masking patterns perform a similar arc-shaped trend after fine-tuning.}
   \label{fig:m_patt}
\end{figure}

\subsection{Experimental Settings}
\textbf{Datasets.} Following a standard SSL evaluation recipe~\cite{2021arXiv211106377H, 2022arXiv220503892G}, we conduct end-to-end fine-tuning or linear probing for classification on ImageNet-1K and transfer learning for object detection/instance segmentation on COCO~\cite{TsungYiLin2014MicrosoftCC} and semantic segmentation on ADE20k~\cite{Zhou2017ScenePT}. 
Specifically, ImageNet-1K~\cite{Deng2009ImageNetAL} is organized according to the WordNet hierarchy. It has 1000 classes and roughly 1.28M images, which is normally leveraged to evaluate the progress in image recognition across a wider variety of objects. 
MS-COCO 2017~\cite{TsungYiLin2014MicrosoftCC} contains $\sim$118k images for training, 5k for validation, and $\sim$20k for testing. ADE20K, as a common semantic segmentation dataset, contains 20k images of 150 fine-grained categories.

\textbf{Architectures.}
Following prior arts~\cite{2021arXiv211106377H,2021arXiv210608254B,2021arXiv211109886X}, we use ViT~\cite{2021arXiv210313915S} with different scales as backbones, \textit{i.e.}, ViT-B/16 and ViT-L/16, where $/16$ denotes the patch size. We pre-train and fine-tune the model with the image size of $224 \times 224$. 
To further unleash the potential of the proposed DMJD, we extend evaluations to hybrid convolution-transformer architectures, \textit{eg.} ConViT~\cite{2022arXiv220503892G} with a multi-scale decoder.  
For segmentation, we build FPN on the 3rd, 5th, 7th, and 11th layers of output features following UperNet~\cite{Xiao2018UnifiedPP} and resize images to $512 \times 512$. 
The multi-scale features from ConViT are also fed into the MaskRCNN~\cite{He2020MaskR} head for object detection.

\textbf{Hyper-parameters.}
We pre-train models on ImageNet-1K~\cite{Deng2009ImageNetAL} following the settings of MAE~\cite{2021arXiv211106377H}, where the decoder has a depth of 8 and a width of 512. Concretely, our DMJD is optimized by AdamW~\cite{Loshchilov2019DecoupledWD} with a batch size of 1024, a learning rate of 1.5e-4, and a weight decay of 0.05. 
We pre-train our model for 800 epochs and 20 epochs warmup with the cosine decay learning rate schedule~\cite{2016arXiv160803983L}. Only random resize crop and horizontal flipping are employed for data augmentation in the pre-training stage.  
For DM, the number of masked views $K$ is set to 2 since the prediction rate can easily saturate to 1 under a corruption rate larger than 0.5. The block-wise masking with a corruption rate of 0.6 is adopted. 

\textbf{Evaluation Metrics.}
As methods with DM actually have more views with the same pre-training epochs, we adopt the metric of effective training epochs (ETE)~\cite{2021arXiv211107832Z} for fair training efficiency comparisons, where $ETE=K \times Epoch$. The training efficiency is quantified by GPU hours (Gh.) achieving comparable performance on $8\times$ Nvidia A100 (40GB).

\begin{table}[t]
    \centering
    \caption{Ablation study on the adaptive learning rate scale rule.}
    \label{tab:adlrs}
        \begin{tabular}{l c c}
            \toprule
           Learning Rate Scale Rule & $ \eta_{base} \cdot b/256$ & $ \eta_{base} \cdot (b_u \frac{m_{pred}}{m_{corr}})/256$ \\
            \midrule
            Fine-tuning & 82.1 & 82.4 \\
            \bottomrule
        \end{tabular}
\end{table}

\begin{table}[t]
\begin{center}
  \caption{Training efficiency of different masking patterns. \textit{``Gh."} records GPU hours taken to reach the fine-tuning accuracy (``Ft.") of MAE with 1600 epochs (\textit{i.e.}, 83.6 \%).}
  \label{tab:pattern}
    \begin{tabular}{lcccccc}
    \toprule
     Method & $m_{patt}$ & Epoch & ETE & Ft. & Gh. & Speedup \\
      \midrule
     MAE~\cite{2021arXiv211106377H} & Uniform & 1600 & 1600 & 83.6 & 227 & $1\times$ \\
     \quad \textbf{+DM} & Uniform & 800 & 1600 & 83.6 & 127 & $1.8\times$ \\
     \quad \textbf{+DM} & Block-wise & 400 & 800 & 83.6 & 70 & $3.2\times$ \\
      \bottomrule
    \end{tabular}
\end{center}
\end{table}

 \subsection{Ablation Study}
\subsubsection{\textbf{Disjoint Masking}}
We evaluate DM based on MAE~\cite{2021arXiv211106377H} with normalized pixels as reconstruction targets and only pre-train 100 epochs for fast ablation study on ImageNet-1K. The uniform masking is applied with $m_{corr}$ of 0.75. 

\textbf{Learning rate scale.}
As shown in Table~\ref{tab:adlrs}, the learned model achieves 0.3$\%$ fine-tuning classification accuracy gain with the newly proposed learning rate scale rule, which illustrates the efficacy of the modification. 

\textbf{Masking pattern $m_{patt}$.}
We figure out the optimal corruption rates for specific masking patterns with DM in Fig.~\ref{fig:m_patt}. Concretely, block-wise masking~\cite{2021arXiv211109886X} tends to remove large patch blocks and makes the reconstruction more difficult, that it prefers a lower optimal corruption rate around 60$\%$ rather than 70$\%$ in uniform masking.

In Table~\ref{tab:pattern}, we further compare the convergence performances of different masking patterns with sufficient training. Equipped with DM of uniform masking, MAE achieves $1.8\times$ convergence speedup.
Notably, block-wise masking~\cite{2021arXiv211109886X} reaches the same accuracy using only half of the ETEs and $3.2\times$ lower Gh. 
This is because masking patterns with smaller optimal masking rate (\textit{e.g.} block-wise) have more room for prediction rate increment which may take more advantages of DM. These facts clearly justified the training efficiency of our DM on MAE.

\begin{figure*}[!t]
\centering
    \subfloat[Prediction rates of DM.]
    {
        \includegraphics[width = 0.23\textwidth]{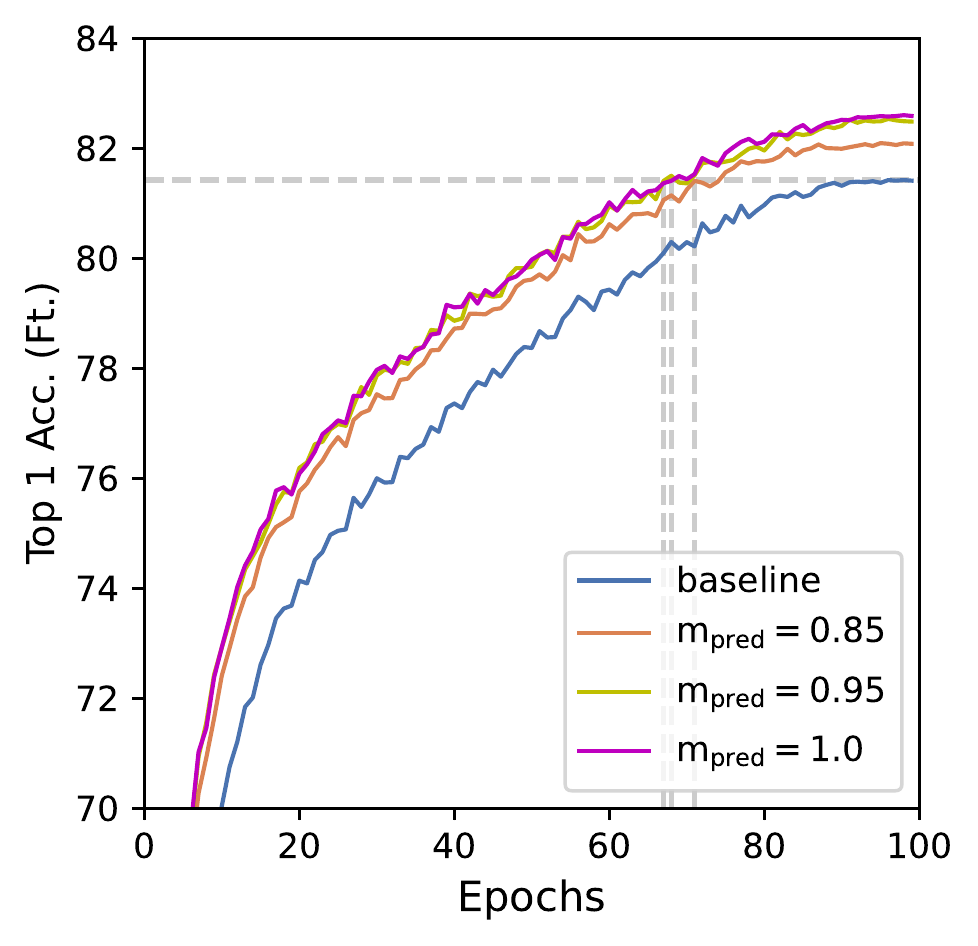}
        \label{fig:pred}
    }
    \hfill
    \subfloat[Learning targets of JD.]
    {
        \includegraphics[width = 0.23\textwidth]{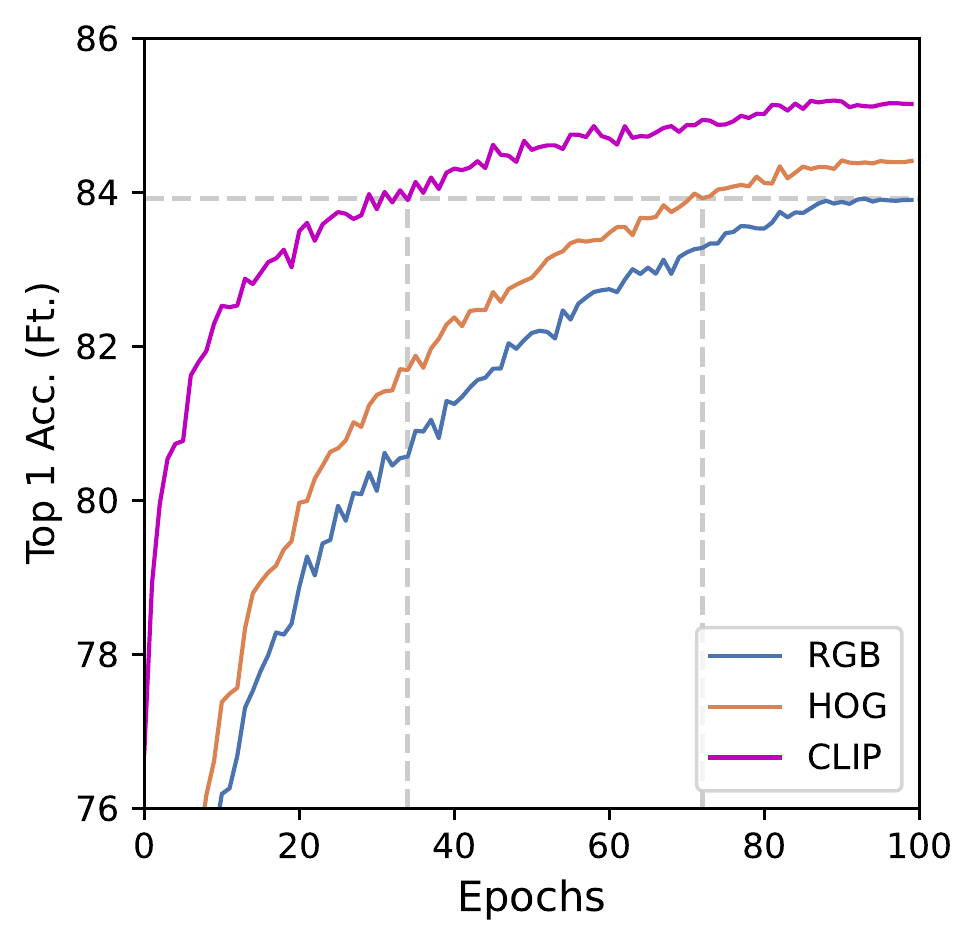}
        \label{fig:tar_e}
    }
	\hfill
	\subfloat[Architecture settings of JD.]
    {
        \includegraphics[width = 0.23\textwidth]{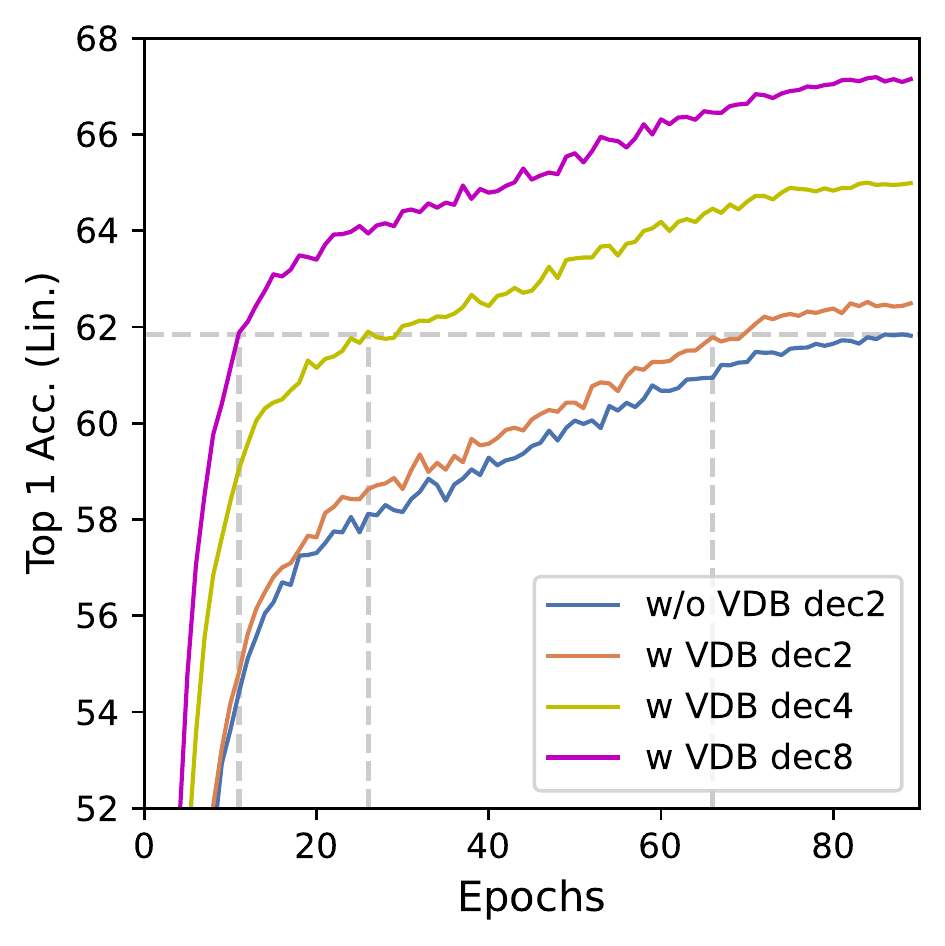}
        \label{fig:vdb_e}
    }
    \hfill
	\subfloat[Main components of DMJD.]
    {
        \includegraphics[width = 0.23\textwidth]{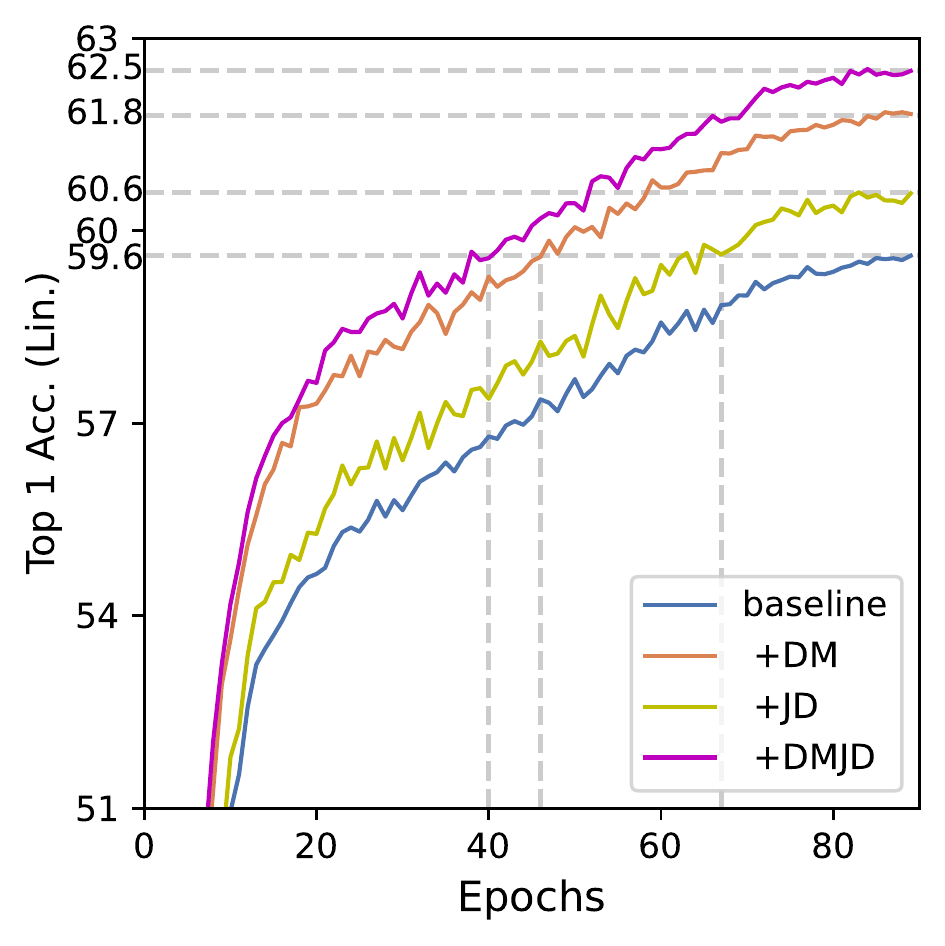}
        \label{fig:dmjd_e}
    }
\caption{Ablation study on training efficiency of our DMJD: (a) higher prediction rates with fixed $m_{corr}$ of 0.75 boosts the model training; (b) learning target types account for convergence speed; (c) MPB with a deep decoder and VDB both benefit learning efficiency; (d) DM and JD work cooperatively.} 
\end{figure*}

\textbf{Prediction rate $m_{pred}$.}
With fixed $m_{corr}$, a larger prediction rate ensures a higher training efficiency which saturates at around 95$\%$, Fig.~\ref{fig:pred}. It justifies that DM facilitates flexible controlling of the prediction rate to make full use of training signals which leads to superior model convergence.

\begin{table}[t]
\begin{center}
  \caption{Training efficiency comparisons with our DM on ImageNet-1K for classification and ADE20K for semantic segmentation (mIoU). $\dagger $ denotes the block-wise masking with a optimal corruption rate of 0.6 is adopted.}
  \label{tab:efficient}
      \begin{tabular}{lccccc}
        \toprule
        Models        & ETE & Gh. & Fine-tuning & mIoU & Speedup \\
        \midrule
        \emph{ViT-Base}       \\
        BEiT~\cite{2021arXiv210608254B} & 800 & 184 & 83.4 & 45.6 & $1\times$ \\
        \rowcolor{mygray} \quad \textbf{+DM} &   600   & 89 & 83.5 & 45.6 & $2.1\times$ \\
        SimMIM~\cite{2021arXiv211109886X} &   800   & 120 & 83.8 & - & $1\times$ \\
        \rowcolor{mygray} \quad \textbf{+DM} &   800   & 86 & 83.8 & - & $1.4\times$ \\
        MAE~\cite{2021arXiv211106377H} &   1600   & 227 & 83.6 & 48.1 & $1\times$ \\
        \rowcolor{mygray} \quad \textbf{+DM} &   1600   & 127 & 83.6 & 48.3 & $1.8\times$ \\
        MaskFeat~\cite{2021arXiv211209133W} &  1600  & 240 & 84.0 & - & $1\times$ \\
        \rowcolor{mygray} \quad $\textbf{+DM}^\dagger$ &  800  & 66 & 84.0 & - & $3.6\times$ \\
        \midrule
        \emph{ViT-Large}      \\
        MAE~\cite{2021arXiv211106377H} &   1600  & 272 & 85.9 & 53.6 & $1\times$ \\
        \rowcolor{mygray} \quad $\textbf{+DM}^\dagger$ & 800 & 100 & 85.9 & 53.5 & $2.7\times$ \\
        MaskFeat~\cite{2021arXiv211209133W} &  1600  & 280 & 85.7 & - & $1\times$ \\
        \rowcolor{mygray} \quad $\textbf{+DM}^\dagger$ &  800  & 76 & 85.9 & - & $3.7\times$ \\
        \bottomrule
      \end{tabular}
\end{center}
\end{table}

\textbf{Efficiency Effects.}
In Table~\ref{tab:efficient}, we evaluate the training efficiency of DM based on several MIM methods with their original pre-training recipes, including MAE~\cite{2021arXiv211106377H}, BEiT~\cite{2021arXiv210608254B}, SimMIM~\cite{2021arXiv211109886X}, and MaskFeat~\cite{2021arXiv211209133W}. 
Specifically, we compare the end-to-end fine-tuning classification accuracy on ImageNet-1K and the transfer learning performance of semantic segmentation (without multi-scale testing) on ADE20K~\cite{Zhou2017ScenePT}. 

On the one hand, DM significantly reduces the training time with equal or fewer ETEs to achieve similar (sometimes better) classification accuracy, \textit{i.e.}, $1.4\times \sim 3.7\times$ speedup. This is because DM requires fewer disk I/O, demonstrating it as a hardware-friendly strategy. 
On the other hand, with block-wise masking, DM further accelerates model convergence. 
For example, MAE-$L$+$DM$ achieves 85.9$\%$ classification accuracy with only 800 ETEs, half of the original budgets. And so does MaskFeat. These facts indicate that our DM remarkably relieves the long pre-training requirement for MIM and accelerates model convergence greatly.

For segmentation, DM also achieves faster convergence and performs on par with the baselines. 
For example, MAE-$L$+$DM$ pre-trained with only 800 ETEs is competitive to the model with 1600 ETEs, speeding $2.7\times$. These results evidence that DM keeps the model generalizability during training acceleration.

\subsubsection{\textbf{Joint Distillation}}
\label{ablate:jd}
We ablate JD based on ConvMAE-B~\cite{2022arXiv220503892G} with DM for faster convergence and pre-train 50 epochs with a batch size of 1024, the learning target as HOG features~\cite{Dalal2005HistogramsOO} normalized by a non-parametric LN layer~\cite{2016arXiv160706450L}. 

\textbf{Learning Targets} are critical for MIM since they provide explicit guidance for representation abstraction and their characteristics can be directly injected into the learned model.
In Table~\ref{tab:vdb-model}, we can conclude that the target normalization is important for good performance, which aligns with the observations in MAE~\cite{2021arXiv211106377H}. Specifically, non-parametric layer normalization~\cite{2016arXiv160706450L} is slightly more suitable for the HOG~\cite{Dalal2005HistogramsOO} target than the original L2 normalization (84.4$\%$ \textit{vs.} 84.2$\%$). With the CLIP target, layer normalization still works well and reports superior accuracy of 85.2 $\%$.

\begin{table}[!t]
    \centering
    \caption{Ablation study of learning targets in the proposed JD.} 
    \label{tab:vdb-model}
    \begin{tabular}{lccccc}
      \toprule
      Learning Targets & RGB & RGB & HOG & HOG & CLIP\\
      \midrule
      $norm(\cdot)$ & $\times$ & $\checkmark$ & $\checkmark$ & LayerNorm & LayerNorm \\
      Fine-tuning & 83.1 & 83.9 & 84.2 & 84.4 & 85.2 \\
      \bottomrule
    \end{tabular}
\end{table}

\begin{table}[!t]
    \centering
    \caption{Ablation study on architecture settings in JD, \textit{i.e.}, the depth of the decoder in MPB and the projector in VDB. ``-" denotes VDB is not used.}
    \label{tab:vdb}
    \begin{tabular}{lccccc}
      \toprule
      Decoder Depth & 2 & 2 & 2 & 4 & 8 \\ 
      Projector  & - & Linear & Nonlinear & Nonlinear & Nonlinear \\ 
      \midrule
      Linear Probing & 61.8 & 60.4 & 62.5 & 65.0 & 67.2 \\
      \bottomrule
    \end{tabular}
\end{table}

\begin{table*}
\begin{center}
  \caption{Performance comparisons with state-of-the-art SSL methods on ImageNet-1K. The extra ETEs for pre-training tokenizers are formulated by normalized ImageNet-1k epochs~\cite{2021arXiv211209133W}. GPU hours of methods that do not release their code are skipped by ``-". Models trained with huge effective epochs are de-emphasized with light grey (Ditto for other tables).}
  \label{tab:sota}
      \begin{tabular}{lccclcll}
        \toprule
        Method    & Publication & Backbone & ETE & Gh. & Learning Target & Fine-tuning & Linear Probing \\
        \midrule
        \emph{non-parametric tokenizer}       \\
        SplitMask~\cite{2021arXiv211210740E} & Arxiv2021 & ViT-B & 600 & - & Random Patches & 83.6 & 46.5 \\
        MAE~\cite{2021arXiv211106377H} & CVPR22 & ViT-B & 1600 & 227 & RGB & 83.6 & 68.0 \\
        SimMIM~\cite{2021arXiv211109886X} & CVPR22 & ViT-B & 800 & 184 & RGB & 83.8 & 56.7 \\
        MaskFeat~\cite{2021arXiv211209133W} & CVPR22 & ViT-B & 1600 & 240 & HOG & 84.0 & 68.5 \\
        \rowcolor{mygray} \quad \textbf{+DMJD} & -& ViT-B & 1600 & 132 \textcolor{Green}{(1.8$\times$)} & HOG & 84.1 \textcolor{Green}{(+0.1)} & 71.9 \textcolor{Green}{(+3.4)} \\
        ConvMAE~\cite{2022arXiv220503892G} & NeurIPS22 & ConViT-B & 1600 & 300 & RGB & 85.0 & 70.9 \\
        \rowcolor{mygray} \quad \textbf{+DMJD} &- & ConViT-B & 800 & 101 \textcolor{Green}{(3$\times$)} & HOG & 85.2 \textcolor{Green}{(+0.2)} & 76.7 \textcolor{Green}{(+5.8)} \\
        ConvMAE~\cite{2022arXiv220503892G} & NeurIPS22 & ConViT-L & 800 & 480 & RGB & 86.2 & - \\
        \rowcolor{mygray} \quad \textbf{+DMJD} &- & ConViT-L & 800 & 267 \textcolor{Green}{(1.8$\times$)} & HOG & \textbf{86.3} \textcolor{Green}{(+0.1)} & \textbf{79.7} \\
        \hline
        \emph{parametric tokenizer}      \\
        MoCov3~\cite{chen2021empirical} & ICCV21 & ViT-B & 1200 & - & Momentum & 83.2 & 76.7 \\
        DINO~\cite{caron2021emerging} & ICCV21 & ViT-B & 1600 & - & Momentum & 83.6 & 78.2 \\
        BEiT~\cite{2021arXiv210608254B} & ICLR22 & ViT-B & 800+1199 & - & DALL-E & 83.4 & 37.6 \\
        CAE~\cite{2022arXiv220203026C} & Arxiv2022 & ViT-B & 1600+1199 & - & DALL-E & 83.9 & 70.4 \\
        PeCo~\cite{2021arXiv211112710D} & CVPR22 & ViT-B & 800+100 & - & Perceptual Codebook & \textbf{84.5} & - \\
        mc-BEiT~\cite{Li2022mcBEiTMD} & ECCV22 & ViT-B & 800+1199 & - & Perceptual Codebook & 84.1 & - \\
        iBOT~\cite{2021arXiv211107832Z} & ICLR22 & ViT-B & 1600 & 233 & Momentum & 84.0 & \textbf{79.5} \\
        SdAE~\cite{2022arXiv220800449C} & ECCV22 & ViT-B & 1200 & - & Momentum & 84.1 & 64.9 \\
        data2vec~\cite{Baevski2022data2vecAG} & ICML22 & ViT-B & 800 & - & Momentum & 84.2 & - \\
        MVP~\cite{2022arXiv220305175W} & Arxiv2022 & ViT-B & 300+10000 & - & CLIP & \textcolor{lightgray}{84.4} & \textcolor{lightgray}{75.4} \\
        ConvMAE~\cite{2022arXiv220503892G} & NeurIPS22 & ConViT-B & 400+10000 & - & CLIP & \textcolor{lightgray}{85.2} & - \\
        \rowcolor{mygray} \quad \textbf{+DMJD} & -& ConViT-B & 400+10000 & - & CLIP & \textcolor{lightgray}{85.4} \textcolor{Green}{(+0.2)} & \textcolor{lightgray}{80.1} \\
        \rowcolor{mygray} \quad \textbf{+DMJD} & -& ConViT-L & 400+10000 & - & CLIP & \textcolor{lightgray}{86.8} \textcolor{Green}{(+1.6)} & \textcolor{lightgray}{81.0} \\
        \bottomrule
      \end{tabular}
\end{center}
\end{table*}

In Fig.~\ref{fig:tar_e}, we illustrate the effects on training efficiency with different targets and reveal that targets with high-level semantics may achieve superior learning efficiency.
Concretely, compared to HOG, raw pixels come with more sensitivity and ambiguity in the masked prediction with wrong color and texture~\cite{2021arXiv211209133W}. The corresponding high loss penalty will mislead the model to overfit local statistics, which is insignificant for visual understanding, resulting in training inefficiency. As high-level semantic abstractions, targets extracted by CLIP significantly boost the model convergence compared with HOG features and raw pixels.

\textbf{Masked Prediction Branch.}
Since specialized in reconstruction, the decoders in an autoencoder with different depths report sensitive linear probing performances. Specifically, as shown in Table~\ref{tab:vdb} and Fig.~\ref{fig:vdb_e}, compared with a depth of 2, the decoder with a depth of 8 gains 4.7$\%$ linear probing accuracy with higher training efficiency. It is because a deep decoder can reconstruct more low-level details, maintaining the latent representations with more semantic abstraction.

\textbf{Visible Distillation Branch.}
In Table~\ref{tab:vdb}, we study the effect of varying projector configurations in VDB. With the decoder depth of 2, the nonlinear projector setting reports 2.1$\%$ higher accuracy than the linear projector and outperforms the model learned without VDB by 0.7$\%$.
This is because a powerful projector can benefit representation learning by relieving the backbone's burden of target fitting, similar to the depth effect of the decoder in MDP. 
The convergence curves of whether or not VDB is used are plotted in Fig.~\ref{fig:vdb_e}, which comprehensively proves the advantages of our JD for MIM training efficiency.

\subsubsection{\textbf{Disjoint Masking \& Joint Distillation}}
Following the settings in Sec.~\ref{ablate:jd}, we conduct evaluations with ConvMAE to justify the training efficiency effects of each component of our DMJD.
In Fig.~\ref{fig:dmjd_e}, one can see that DM and JD both improve the training efficiency significantly and their combination further  accelerates the model convergence remarkably.
This is because they are devised from orthogonal perspectives and thus work cooperatively. 

\subsection{Performance Comparisons}
In Table~\ref{tab:sota}, we roughly categorize the leading baseline methods by whether or not the supervision signal stems from a parametric tokenizer. 
Non-parametric tokenizers include pixels or hand-crafted features, while parametric ones are deep features from an online or pre-trained teacher network.
For a fair efficiency comparison, the additional cost of pre-training tokenizer, \textit{eg.}  CLIP~\cite{Radford2021LearningTV}, needs to be counted on. 
Following MaskFeat~\cite{2021arXiv211209133W}, we normalize the external training epochs by the cost of each epoch on ImageNet-1K training set with the view size of $224 \times 224$ for unified evaluation.

\begin{figure*}[!t]
\centering
	\subfloat[MAE\_ViT-B~\cite{2021arXiv211106377H}]{\includegraphics[width = 0.24\textwidth]{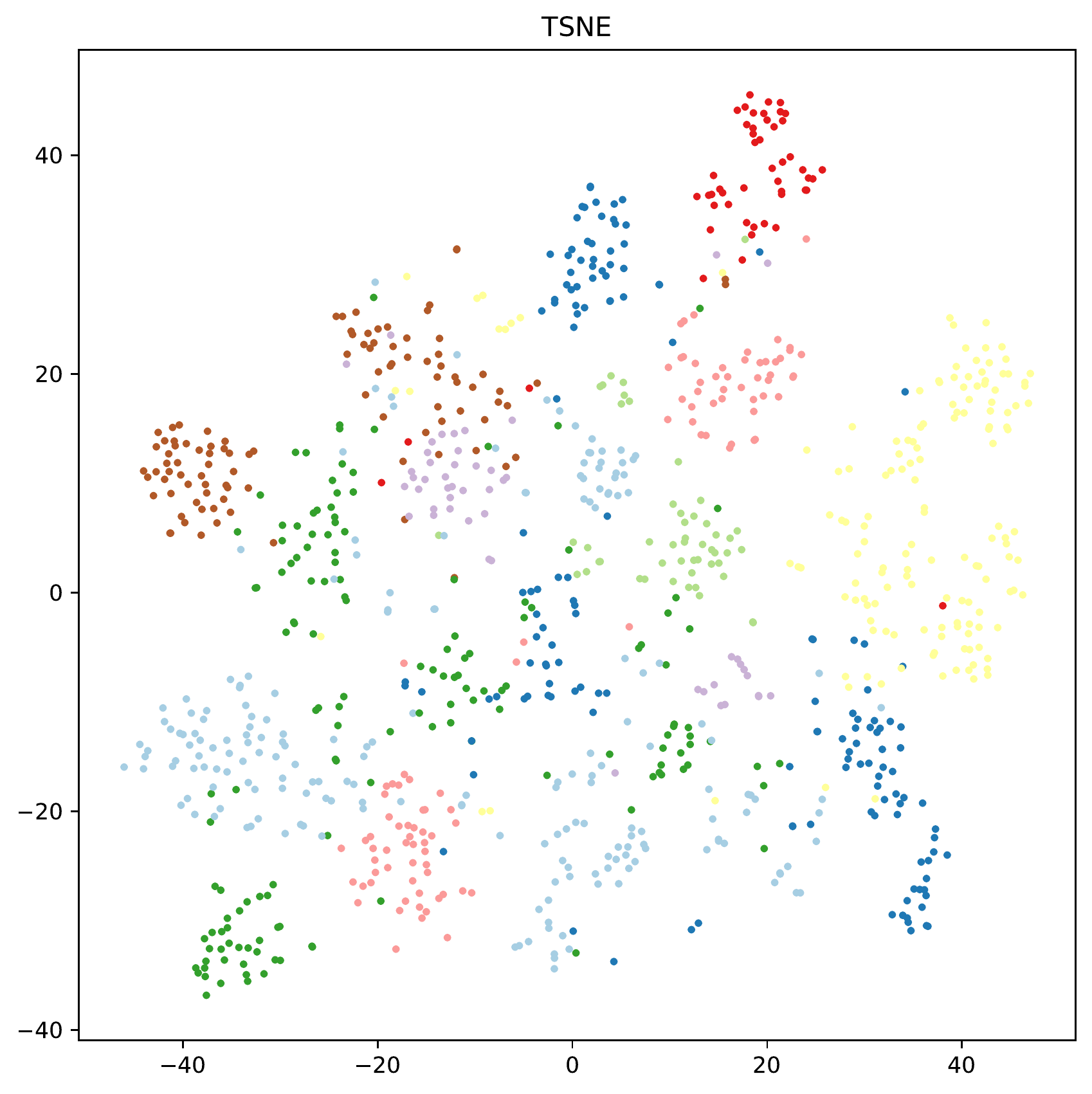}}
	\hfill
    \subfloat[Maskfeat\_ViT-B~\cite{2021arXiv211209133W}]{\includegraphics[width = 0.24\textwidth]{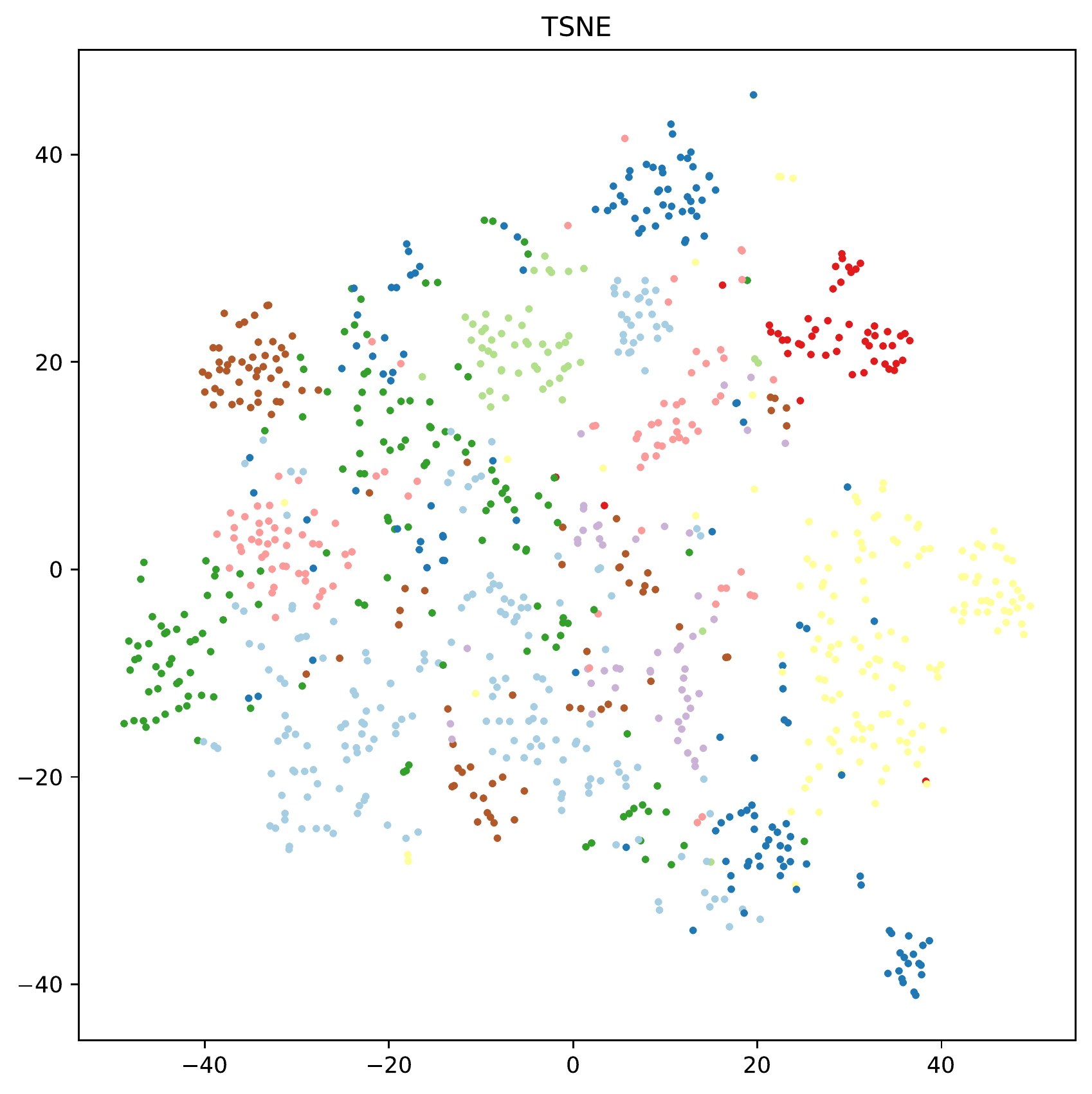}}
	\hfill
    \subfloat[DMJD\_HOG\_ViT-B (\textbf{Ours})]{\includegraphics[width = 0.24\textwidth]{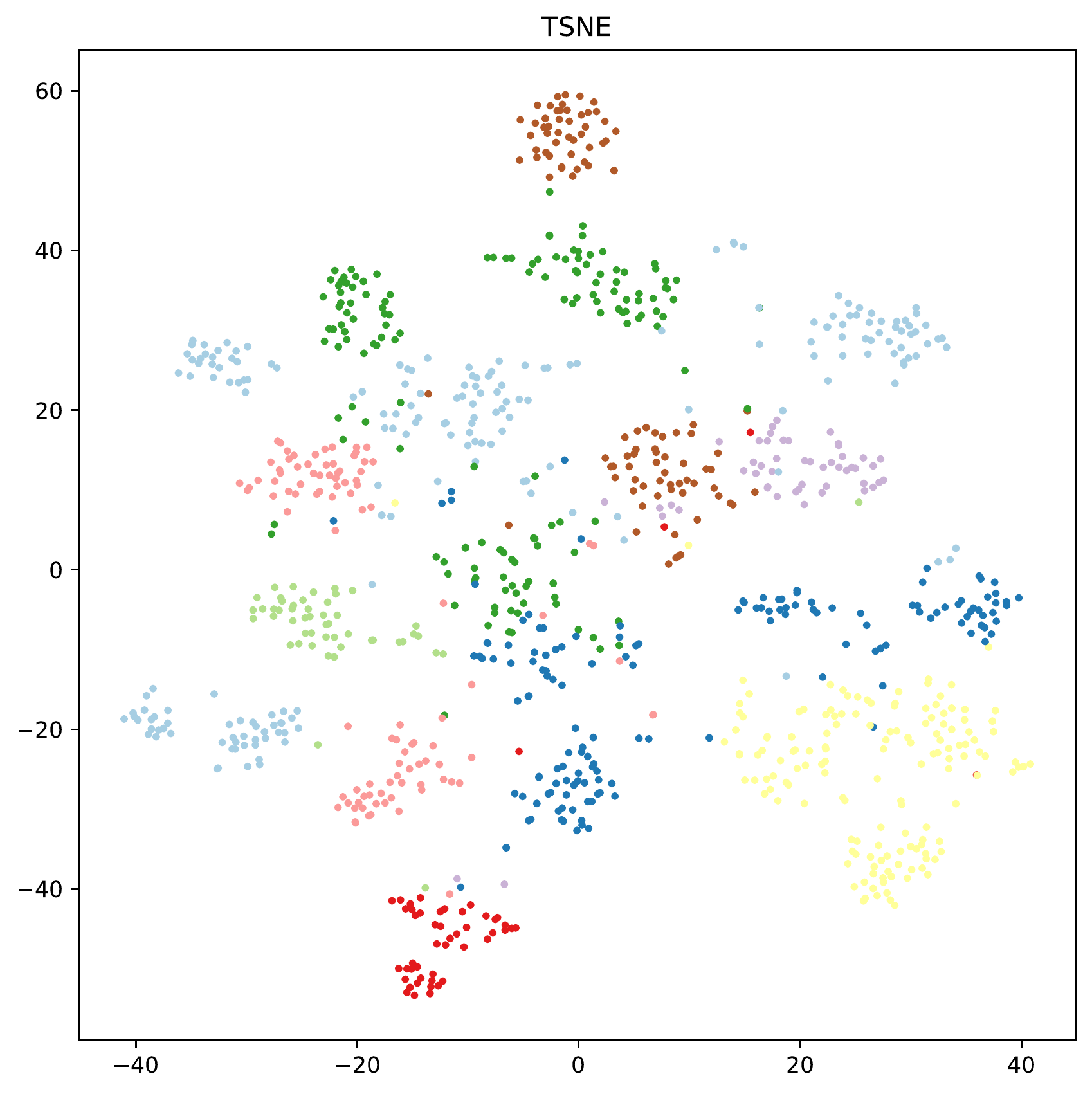}}
    \hfill	
	\subfloat[ConvMAE\_ConViT-B~\cite{2022arXiv220503892G}]{\includegraphics[width = 0.24\textwidth]{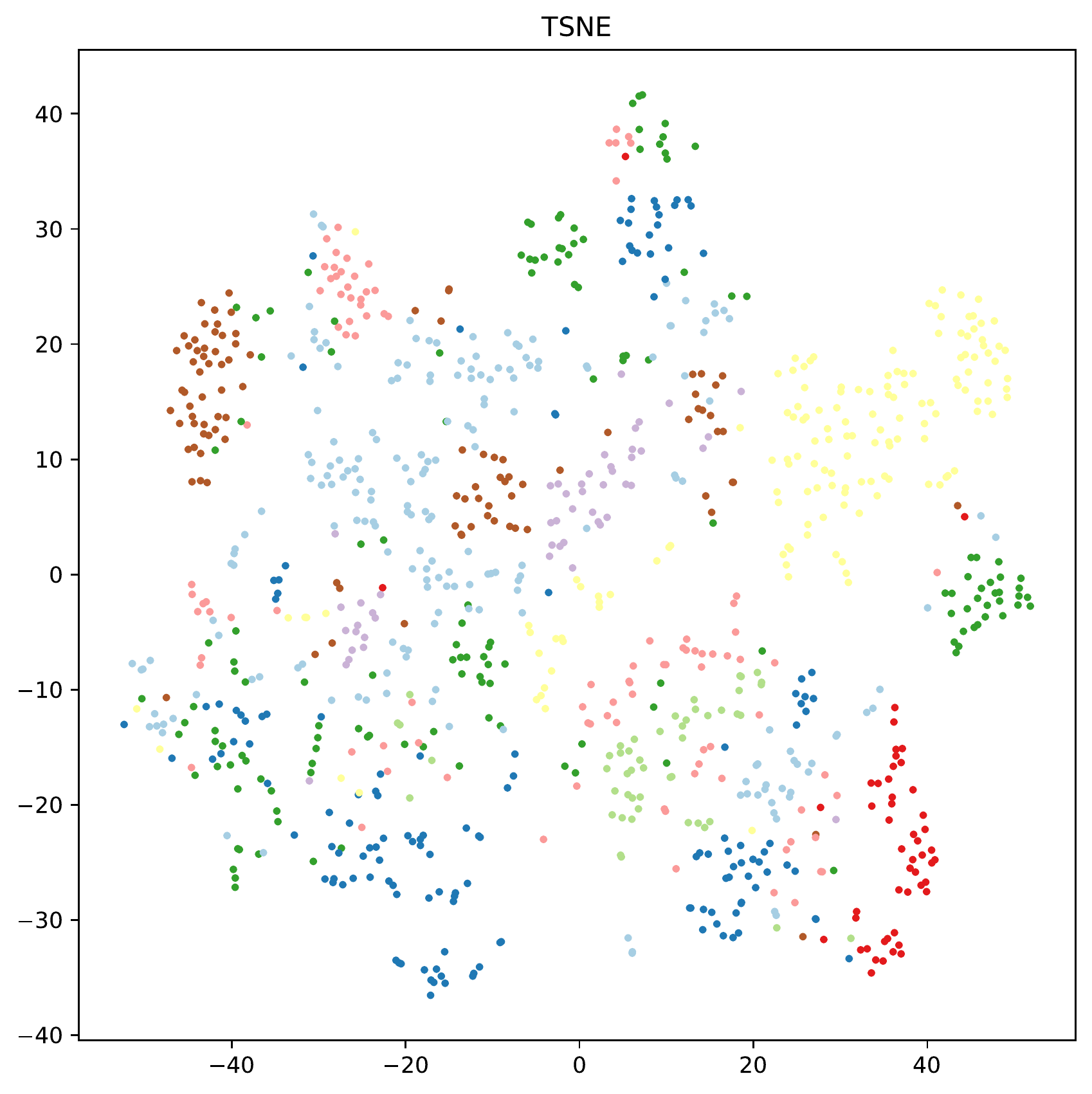}}
    \quad
    \subfloat[DMJD\_HOG\_ConViT-B (\textbf{Ours})]{\includegraphics[width = 0.24\textwidth]{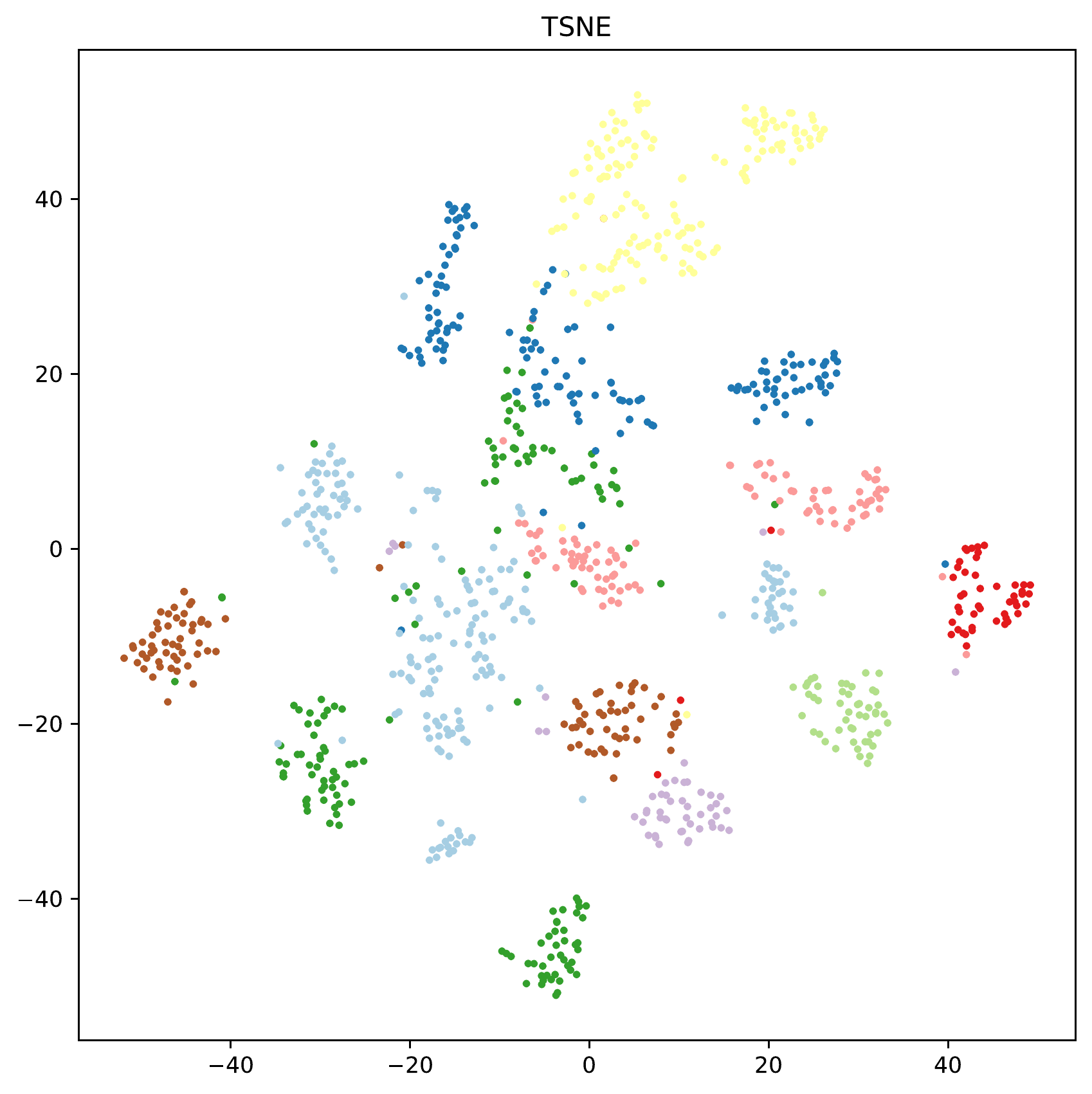}}
	\hfill
    \subfloat[DMJD\_HOG\_ConViT-L (\textbf{Ours})]{\includegraphics[width = 0.24\textwidth]{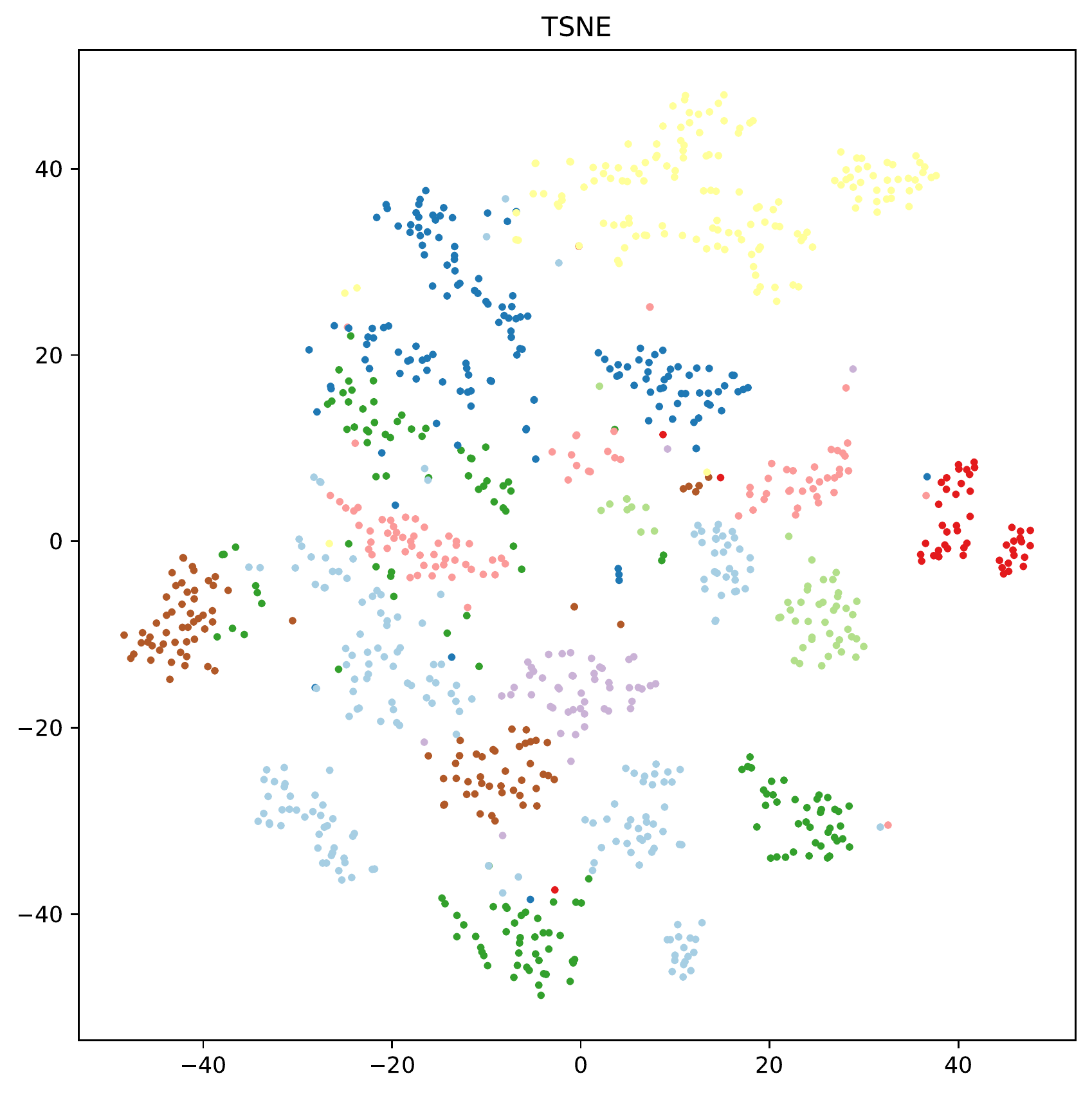}}
	\hfill
    \subfloat[DMJD\_CLIP\_ConViT-B (\textbf{Ours})]{\includegraphics[width = 0.24\textwidth]{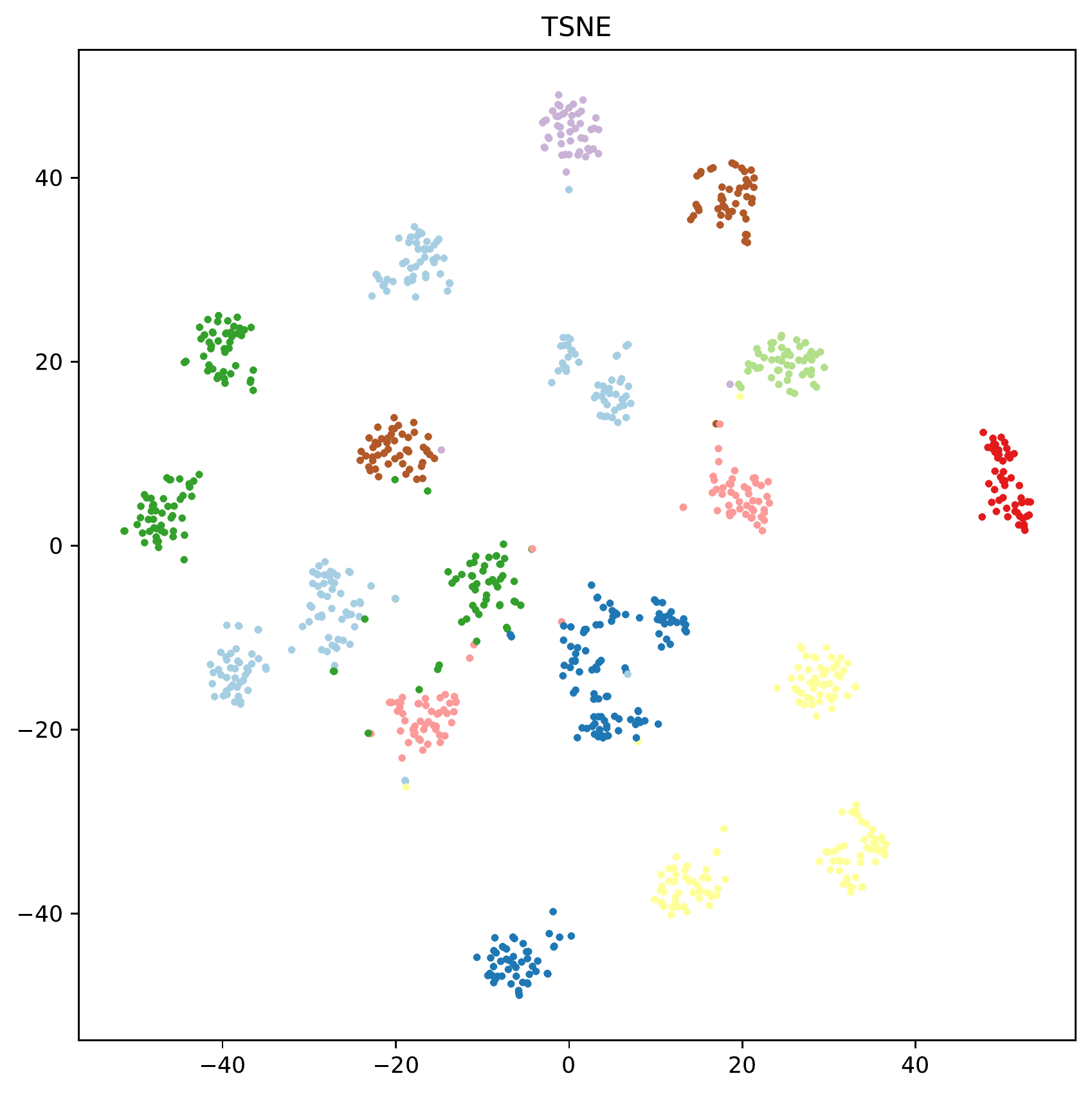}}
    \hfill	
	\subfloat[DMJD\_CLIP\_ConViT-L (\textbf{Ours})]{\includegraphics[width = 0.24\textwidth]{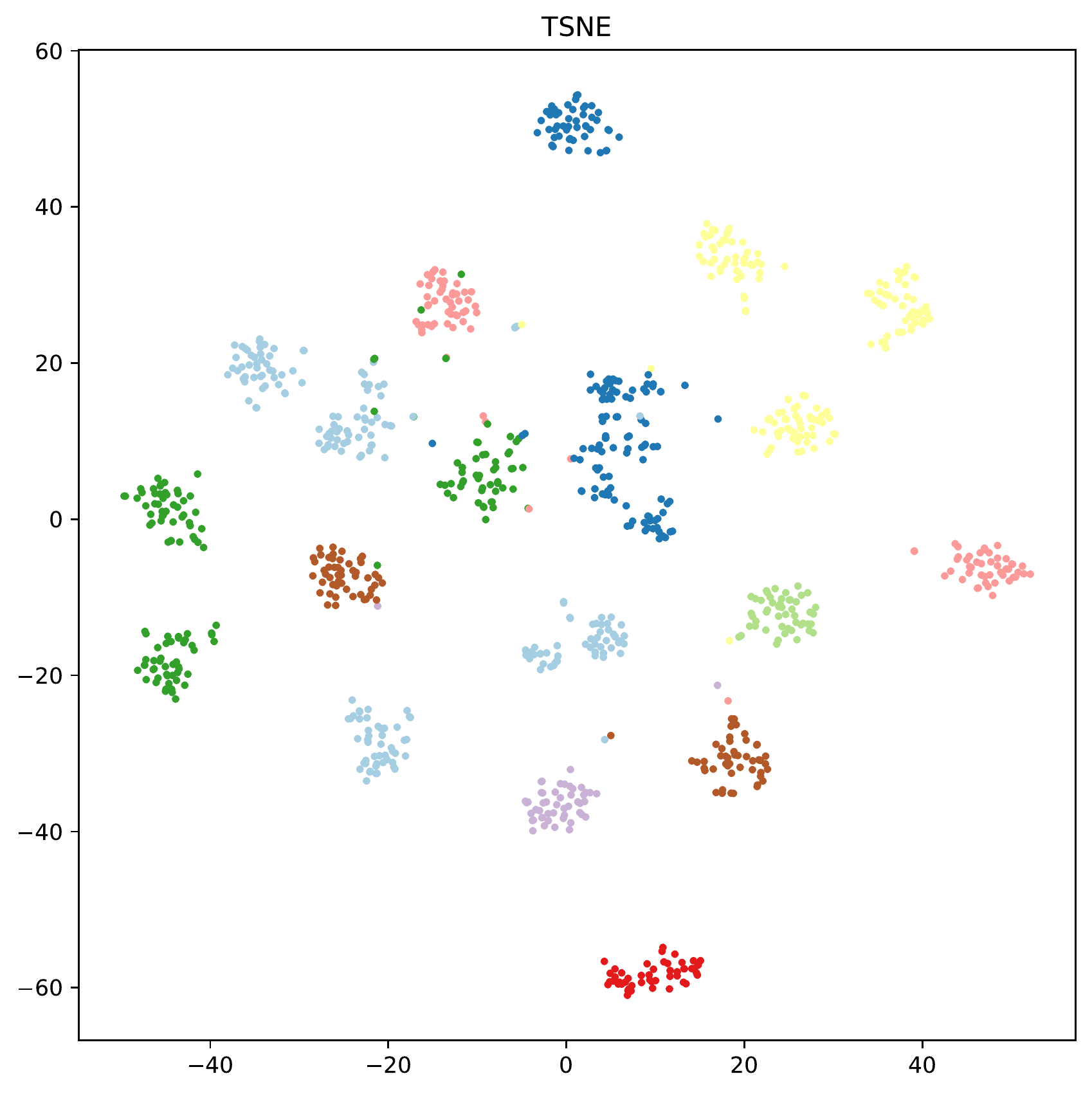}} 
\caption{t-SNE visualization of representation spaces learned by state-of-the-art methods (fully pre-trained with ETEs of 1600, 1600, 1600, 1600, 800, 800, 400, and 400 respectively), where test images from 20 classes are randomly sampled in the ImageNet-1K validation set.}
\label{fig:tsne}
\end{figure*}

\begin{table}[!t]
  \centering
  \caption{Performance comparisons of semantic segmentation on ADE20K (mIoU) and object detection and instance segmentation on COCO (AP$^b/$AP$^m$) with leading MIM methods.}
  \label{tab:det}
      \begin{tabular}{lcccl}
        \toprule
        Method        & Backbone & ETE  & mIoU & AP$^b/$AP$^m$ \\
        \midrule
        \emph{non-parametric $tkn(\cdot)$}       \\
        SplitMask~\cite{2021arXiv211210740E} & ViT-B & 600 & 45.7 & 46.8$/$42.1 \\
        MAE~\cite{2021arXiv211106377H} & ViT-B & 1600 & 48.1 & 50.3$/$44.9 \\
        SimMIM~\cite{2021arXiv211109886X} & Swin-B & 800 & 52.8 & 52.3$/$- \\
        ConvMAE~\cite{2022arXiv220503892G} & ConViT-B & 1600 & 51.7 & 53.2$/$47.1 \\
        \rowcolor{mygray} \quad \textbf{+DMJD}  & ConViT-B & 800 & \textbf{53.3} & \textbf{53.4$/$47.6} \\
        \hline
        \emph{parametric $tkn(\cdot)$}      \\
        DINO~\cite{caron2021emerging} & ViT-B & 1600 & 46.8 & 50.1$/$43.4 \\
        BEiT~\cite{2021arXiv210608254B} & ViT-B & 1999 & 47.1 & 50.1$/$43.5 \\
        mc-BEiT~\cite{Li2022mcBEiTMD} & ViT-B & 1999 & 50.8 & 49.2$/$44.0 \\
        SdAE~\cite{2022arXiv220800449C} & ViT-B & 1200 & 48.6 & 48.9$/$43.0 \\
        PeCo~\cite{2021arXiv211112710D} & ViT-B & 900 & 48.5 & 44.9$/$40.4 \\
        CAE~\cite{2022arXiv220203026C} & ViT-B & 2799 & 50.2 & 50.0$/$44.0 \\
        iBOT~\cite{2021arXiv211107832Z} & ViT-B & 1600 & 50.0 & 51.2$/$44.2 \\
        ConvMAE~\cite{2022arXiv220503892G} & ConViT-B & 10400 & \textcolor{lightgray}{52.8} & \textcolor{lightgray}{53.3$/$47.3} \\
        \bottomrule
      \end{tabular}
\end{table}

\textbf{Classification.}
With a non-parametric tokenizer of HOG, our DMJD improves the fine-tuning and linear probing accuracy by 0.2$\%$, 5.8$\%$ respectively, compared with ConvMAE (ConViT-B) of the RGB target and notably consumes only half of training ETEs (800ep \textit{vs.} 1600ep) and $\frac{1}{3}$ Gh. (300 \textit{vs.} 101). 
With the same targets and a ViT-B backbone, our DMJD consistently improves the linear probing performance with MaskFeat by a large margin (71.9 \textit{vs.} 68.5) with less training cost of Gh.
Overall, it is clear that our DMJD framework can significantly improve the linear probing results in a plug-and-play fashion while decreasing training costs. 

As MIM is obsessed with patch-level semantic relation reasoning, it is typically inferior to contrastive learning-based pre-training models under the linear probing evaluation, where image-level discriminative representations are critical.
However, our DMJD with ConvMAE (ConViT-B) improves the baseline of 5.8\% and achieves comparable performance with state-of-the-art contrastive learning-based SSL methods, e.g., iBOT~\cite{2021arXiv211107832Z} (76.7 \textit{vs.} 79.5). When scaling up the model size from ConViT-B to ConViT-L, our DMJD further reports a higher liner probing accuracy of 79.7.
With CLIP targets, DMJD further pushes the accuracy to another height of 81.0 with ConVit-L.

In Fig.~\ref{fig:tsne}, we present qualitative evaluations of the learned models' representations. The t-SNE visualization presents a trend consistent with the linear probing performance that our DMJD with high-level learning targets produces more discriminative embedding spaces, which again justifies the superiority of our designs.

\textbf{Semantic Segmentation and Object Detection.} 
We conduct fine-grained transfer learning evaluations to verify the generalization ability of learned representations by the proposed DMJD, Table~\ref{tab:det}. Since multi-scale features benefit downstream tasks, we adopt UperNet~\cite{Xiao2018UnifiedPP} and Mask R-CNN~\cite{He2020MaskR} as segmentation and object detection heads, respectively.

Concretely, with a non-parametric target tokenizer, DMJD surpasses the ConvMAE baseline by 1.6 mIoU (53.3 vs. 51.7) on ADE20K for semantic segmentation and achieves 0.2 $AP^{box}$ (53.4 \textit{vs.} 53.2) and 0.5 $AP^{mask}$ (47.6 \textit{vs.} 47.1) gains on COCO for object detection and instance segmentation. 
It is worth noting that our DMJD outperforms models learned with parametric target tokenizers, \textit{eg.} CLIP~\cite{Radford2021LearningTV}, and only consumes half of ETEs.
These results justify that the learned representations of DMJD generalize well for several downstream tasks while keeping efficiency.

\section{Conclusion}
In this paper, we propose a conceptually simple yet training-efficient MIM framework, termed disjoint masking with joint distillation (DMJD). 
With the multi-view disjoint masking strategy, our approach improves the utilization of training signals accelerating the model convergence significantly. 
Introducing explicit semantic guidance on the visible parts via joint distillation, the proposed DMJD cooperatively boosts the training efficiency while keeping the learned representations to generalize well to downstream tasks.
We hope these achievements shed new light on training efficiency research in the MIM community. 

\bibliographystyle{IEEEtran}
\bibliography{reference}

\end{document}